\documentclass[lettersize,journal]{IEEEtran}
\usepackage{amsmath,amsfonts}
\usepackage{algorithm}
\usepackage{algpseudocode}
\usepackage{array}
\usepackage[caption=false,font=normalsize,labelfont=sf,textfont=sf]{subfig}
\usepackage{textcomp}
\usepackage{stfloats}
\usepackage{url}
\usepackage{verbatim}
\usepackage{graphicx}
\usepackage{cite}
\usepackage{multirow}
\usepackage{utfsym}
\begin{document}

\title{TcGAN: Semantic-Aware and Structure-Preserved GANs with Individual Vision Transformer for Fast Arbitrary One-Shot Image Generation}

\author{Yunliang Jiang, Lili Yan, Xiongtao Zhang, Yong Liu, Danfeng Sun	
\thanks{This work was supported in part by National Natural Science Foundation of China (U22A20102), and the “Pioneer” and “Leading Goose” R \& D Program of Zhejiang Province (2023C01150)(\textit{Corresponding author: Xiongtao Zhang}).}
\thanks{Yunliang Jiang is with the School of Computer Science and Technology, Zhejiang Normal University, Jinhua 321004, China, the Zhejiang Province Key Laboratory of Smart Management \& Application of Modern Agricultural Resources, Huzhou University, Huzhou 313000, China, and also with the School of Information Engineering, Huzhou University, Huzhou 313000, China (e-mail: 1047897965@qq.com).} 
\thanks{Lili Yan and Xiongtao Zhang are with the Zhejiang Province Key Laboratory of Smart Management \& Application of Modern Agricultural Resources, Huzhou University, Huzhou 313000, China, and also with the School of Information Engineering, Huzhou University, Huzhou 313000, China.}
\thanks{Yong Liu is with the College of Control Science and Engineering, Zhejiang University, Hangzhou 310027, China.}
\thanks{Danfeng Sun is with the School of Computer Science, Hangzhou Dianzi University, Hangzhou 310018, China.}

}

\markboth{Journal of \LaTeX\ Class Files,~Vol.~14, No.~8, August~2021}%
{Shell \MakeLowercase{\textit{et al.}}: A Sample Article Using IEEEtran.cls for IEEE Journals}


\maketitle

\begin{abstract}
One-shot image generation (OSG) with generative adversarial networks that learn from the internal patches of a given image has attracted world wide attention. In recent studies, scholars have primarily focused on extracting features of images from probabilistically distributed inputs with pure convolutional neural networks (CNNs). However, it is quite difficult for CNNs with limited receptive domain to extract and maintain the global structural information. Therefore, in this paper, we propose a novel structure-preserved method TcGAN with individual vision transformer to overcome the shortcomings of the existing one-shot image generation methods. Specifically, TcGAN preserves global structure of an image during training to be compatible with local details while maintaining the integrity of semantic-aware information by exploiting the powerful long-range dependencies modeling capability of the transformer. We also propose a new scaling formula having scale-invariance during the calculation period, which effectively improves the generated image quality of the OSG model on image super-resolution tasks. We present the design of the TcGAN converter framework, comprehensive experimental as well as ablation studies demonstrating the ability of TcGAN to achieve arbitrary image generation with the fastest running time. Lastly, TcGAN achieves the most excellent performance in terms of applying it to other image processing tasks, e.g., super-resolution as well as image harmonization, the results further prove its superiority.
\end{abstract}

\begin{IEEEkeywords}
One-shot image generation, generative adversarial networks, vision transformer, convolutional neural networks
\end{IEEEkeywords}

\section{Introduction}
\IEEEPARstart{B}{enefited} from the great success of Generative Adversarial Networks (GANs)\cite{goodfellow2014generative3}, researchers have made much progress on computer vision fields\cite{4_1,4_2}, such as style transfer\cite{deng2022stytr24}, image restoration\cite{8_0, liang2021swinir5}, deraining\cite{8}, dehazing\cite{8_1,8_2,8_4}, super-resolution\cite{8_3, ledig2017photo6}, speech synthesis\cite{lee2021multi7}, and video generation\cite{menapace2021playable8}. GANs are trained in an adversarial manner improving the discriminator's ability to identify true and false, as much the quality of the images generated by the generator. 
Nevertheless, conventional GANs are usually based on sufficient data samples for training the deep neural networks, which are time-expensive and tedious in some specific scenarios to collect and label, e.g., medical diagnosis and industrial defects. Therefore, it is quite desired to explore training GANs with a small amount of sample data or a single image.

\begin{figure}[!t]
	\centering
	\includegraphics[width=3.5in]{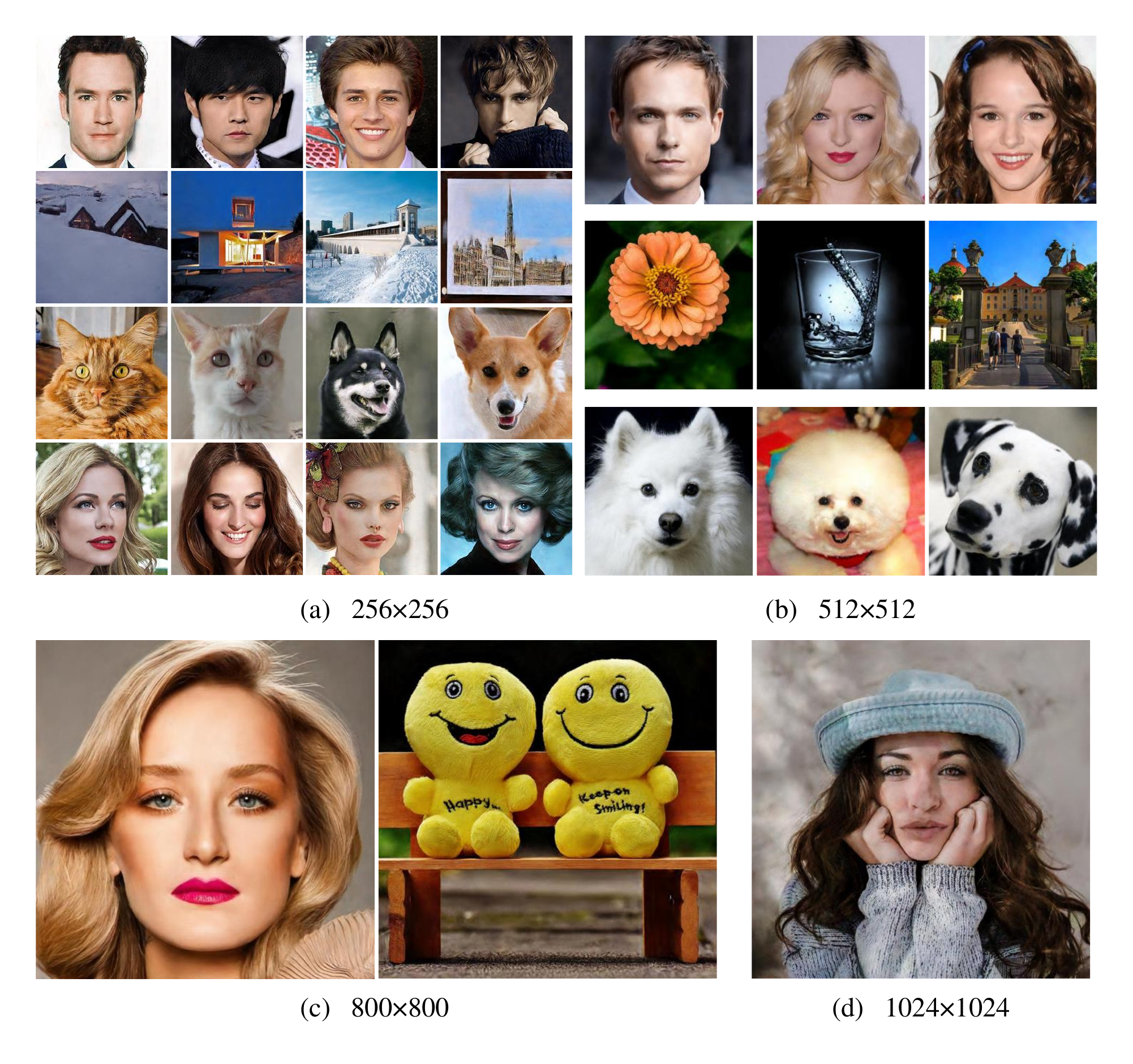}
	\caption{{\bf{Random image samples.}} TcGAN trained on a single image generates realistic random image samples that describe the new global structure and texture details while preserving the data distribution of the training image.}
	\label{fig0}
\end{figure}
Image generation based on single sample training can be traced back to the seminal work SinGAN proposed by Shaham \textit{et al.}\cite{shaham2019singan9}, which demonstrated that training a pure convolutional neural network (CNN) architecture \cite{lecun1989backpropagation10} requires only one image. This task referred to as One-shot image Generation (OSG)\cite{14_1}, achieved impressive success by using multi-stage GANs and multi-scale approach to learn the internal distribution of patches within the image. However, current OSG methods still have a lot of room for improvement in terms of generation quality and training time. Firstly, they propose a universal pure CNN-based generator architecture. CNNs are prevalent in computer vision tasks due to their natural localization and scale invariance. However, it has some limitations in extracting global features. Secondly, the lack of high-level semantic information and explicit global structure representation prevented them from processing object images as well as natural images. Therefore, to address the above-mentioned shortcomings, in this work, we attempt to accomplish OSG with individual structure-preserved vision transformer\cite{vaswani2017attention11}, which meets the requirements of image generation for capturing the global context by exploiting its powerful ability to model long-range dependencies.

There has been evidence that transformer\cite{devlin-etal-2019-bert12,brown2020language13} has made a dramatic leap in the field of natural language processing (NLP) due to its multi-headed self-attentive mechanism. This has directly stimulated the interest of researchers in the field of computer vision since it\cite{chu2021twins15} also require long-range dependencies inherited from transformer to improve the importance of global structural information. Dosovitskiy {\it{et al.}}\cite{dosovitskiy2020image16} proposed Vision Transformer (ViT), which was the first computer vision method based on a pure transformer architecture, and achieved impressive classification peformance. ViT and its variants\cite{chen2021crossvit17},\cite{tolstikhin2021mlp18} were also extended to various tasks such as object detection\cite{Sun_2021_ICCV19}, video recognition\cite{bertasius2021space23}, and image generation\cite{lee2021vitgan24, jiang2021transgan25}, etc. Among them, Lee \textit{et al.}\cite{lee2021vitgan24} proposed ViTGAN which applies transformer to GANs as a way to implement image generation tasks. Jiang \textit{et al.}\cite{jiang2021transgan25} proposed TransGAN based on a pure transformer generator and discriminator architecture. He argued that it requires more data in the training process than CNN, so increasing the training samples with data augmentation can improve the effectiveness of TransGAN. Although ViTGAN\cite{lee2021vitgan24}, TransGAN\cite{jiang2021transgan25} provide us with more options for implementing image generation tasks, these methods still have the following shortcomings: (i) They require large datasets\cite{deng2009imagenet26} to train the proposed method which resulting in high time complexity as well as prohibitive hardware equipment. Moreover, large datasets in real-life are extremely limited. (ii) The use of pure transformer architecture leads to limited size of the generated images which makes it difficult to implement super-resolution applications in the absence of any other operations.

Address the issues above in OSG and inspired by Vision Transformer, in this paper, we propose a novel Semantic-Aware and Structure-Preserved GANs for fast One-Shot Image Generation named as TcGAN. Not only do we study how to learn the high-level semantic information represented by the global structure in the OSG method, but we investigate how to generate real images with a shorter training time. As a result, TcGAN has the following advantages: (i) It can process arbitrary images not limited to natural images with complex structures and textures, as well as objects images with global structural consistency. (ii) The generated images have high-level semantic-aware information with obvious global structure and no blurring, illusions, or overlaps. (iii) The training time is effectively reduced due to the structure of the proposed TcGAN.

Our contributions are summarized as follows:
\begin{itemize}
	\item{We propose a novel one-shot image generation method TcGAN with individual structure-preserved vision transformer, which can resolve the problem of existing methods lacking detailed information about the global structure of object images and pixel blurring with large-scale images. TcGAN learns global as well as local information simultaneously.}
	\item{We design an unsupervised image generation method for super-resolution tasks. The structural features from the global network, along with the proposed scaling method that has scale-invariance allow the generated super-resolution images to retain the original input semantic-aware information without any additional labeling.}
	\item{Qualitative and quantitative experimental results on a large number of datasets demonstrate that TcGAN significantly outperforms state-of-the-art OSG methods in terms of image quality and training speed. In addition, TcGAN can be extended to other relevant scenarios, such as super-resolution and image harmonization, and relevant experimental results also demonstrate the effectiveness of the proposed method.}
\end{itemize}

\section{Related Work}
\subsection{One-shot image Generation(OSG)}
The goal of OSG methods is to obtain a generative model by modeling the internal patch distribution of a single image, escaping from the limitation for large datasets. InGAN proposed by Shocher \textit{et al.}\cite{InGAN27} was the first conditional OSG method based on Encoder and Decoder structure, which can generate a large amount of natural images of different sizes. In contrast, SinGAN\cite{shaham2019singan9} was the first unconditional OSG method with all its generators based on a multi-stage pyramid structure of CNN for generating images of different sizes.
ConSinGAN\cite{hinz2021improved28} improved the training strategy of SinGAN by utilizing in parallel training way to shorten the number of training stages and reduce the training time. Mogan\cite{chen2021mogan29} proposed to divide the image into ROI (Region of Interest) and residual region, and finally output the results by learning and merging them separately. One shot GAN\cite{sushko2021one30} designed a novel discriminator which consisting of two branches for content and layout respectively while adopting a diversity regularization strategy in the training process of the generator. ExSinGAN\cite{zhang2021exsingan31} divided multiple generators into three specific cascaded parts: structural GAN, semantic GAN, and texture GAN make the method interpretable. In addition, the new GAN inversion technique\cite{pan2021exploiting32} was taken in the network structure to improve the performance of the generated image. GPNN\cite{granot2022drop33} proposed a new non-training method that can quickly generate new images by combining the ideas of SinGAN and PatchGAN\cite{barnes2009patchmatch34}. PetsGAN\cite{zhang2022petsgan35} proposed to incorporate both internal and external priors in the training process to overcome the time-expensive and poor image quality problems of SinGAN. SinDiffusion\cite{39_1} proposed to accomplish OSG using a diffusion model\cite{39_2}. Although OSG methods have achieved rapid development, they are limited in terms of semantics and structure since only internal patches of images are acquired for training compared to methods trained on large-scale datasets.
\subsection{Vision Transformer(ViT)}
The transformer\cite{vaswani2017attention11} in the field of NLP can be used to capture long-range dependencies between words by stacking multi-headed self-attention and feed-forward neural networks. This is because that the attention mechanism characterizes the dependencies between any two remote tokens. The introduction of ViT\cite{dosovitskiy2020image16} brings the transformer to a whole new perspective for computer vision, where it obtains embedding patches (similar to tokens in NLP) by splitting the input image into fixed-size patches, linearly embedding, and finally passes through transformer encoder layers for image classification tasks. However, the transformer needs to be pre-trained on a large dataset. Afterward, DeiT\cite{touvron2021training36} proposed to improve the sample efficiency of ViT by using knowledge distillation\cite{hinton2015distilling37} and regularization tricks. 
CrossViT\cite{chen2021crossvit17} proposed a dual-brand ViT and designed an efficient token fusion strategy to extract multi-scale features for implementing the image classification task. PVT\cite{wang2021pyramid38} proposed to introduce a pyramid structure similar to CNN based on ViT, making PVT can be applied as a backbone in image segmentation and detection tasks. Perceiver\cite{jaegle2021perceiver40} proposed to add an attentional tight latent bottleneck consisting of latent units to the transformer encoder to solve the problem of quadratic scaling of the transformer's full attention. T2TViT\cite{yuan2021tokens41} proposed to enhance local information by aggregating adjacent tokens into one token and naming it Tokens-to-Token (T2T) module, as well as designing an efficient backbone to reduce redundancy and enhance feature richness. ViTAE\cite{xu2021vitae42} proposed to fuse multiple convolutional layers with different dilation rates into the transformer to improve the prior knowledge of the missing local information in the transformer.
\subsection{ViT for Image Generation}
There has been studies on applying transformers to image generation tasks to capture long-range dependencies benefiting from the multi-headed self-attention mechanism. Child \textit{et al.}\cite{child2019generating43} demoted global attention to sparse attention by the top-k selection operation, whcih effectively retains the most helpful part of attention while also removing other irrelevant information. This approach effectively reduces the training time and spatial complexity of traditional transformers, in addition to allowing it to be extended to other application scenarios, such as long-distance speech, text, and image data. Esser \textit{et al.}\cite{esser2021taming44} proposed to generate high pixel images by combining the respective advantages of the inductive bias validity of CNN and the expressive power of the transformer. But the customized designs limit the versatility of their method to other imaging tasks. ViTGAN\cite{lee2021vitgan24} proposed to construct GANs with ViT to implement image generation tasks, and also proposed new techniques to ensure its training stability and improve its convergence, outperforming networks such as TransGAN\cite{jiang2021transgan25} and StyleGAN2\cite{karras2020analyzing45}. TransGAN\cite{jiang2021transgan25} proposed two pure transformer architectures for the generator and discriminator, respectively, and divided the generator into six stages that are trained in a stepwise manner to gradually grow the size of the generated images. The method completely discards convolution operation to accomplish the image generation task. To the best of our knowledge, no one has yet used individual vision transformer to solve the problem of poor representation of high-level semantic information of global structure that is prevalent in OSG methods, especially when dealing with object images with global consistency. Therefore, based on the above considerations, we propose TcGAN and present it in detail in the following sections.

\section{Method}
\begin{figure*}[!t]
	\centering
	\includegraphics[width=7.16in]{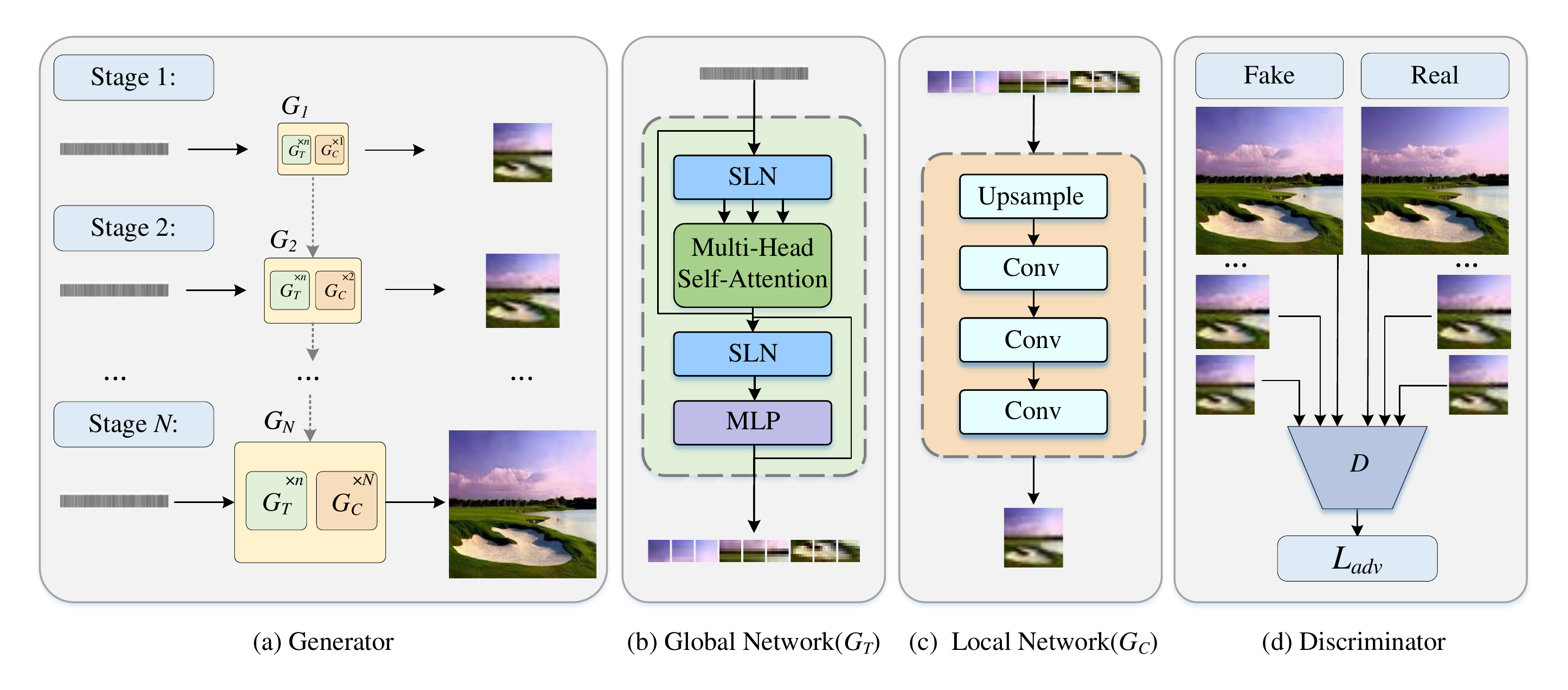}
	\caption{{\bf{Overview.}} The generator of TcGAN consists of two modules, a global network based on the vision transformer architecture and a fully convolutional local network. These two networks use three sets of inputs, a set of random noise for constructing high-level semantic information in the global structure, a global latent variable that defines the image's overall appearance, and noisy images for learning fine-grained texture features. TcGAN synthesizes arbitrary types of images by sequencing global learning followed by local learning.}
	\label{Model}
\end{figure*}
Based on the assumption that any image can be described by both global and local aspects, we propose a stronger OSG method TcGAN according to the designed global and local networks, and Fig. \ref{Model} shows an overview of TcGAN. In this section, we firstly present the overall architecture of TcGAN as shown in Fig. \ref{Model}(a) and the proposed novel scaling method. Then we introduce the detailed design of TcGAN including the global network (denoted as $ G_T $) in Fig. \ref{Model}(b) and local network (denoted as $ G_C $) in Fig. \ref{Model}(c). Finally, the objective function, the learning process of the proposed TcGAN, and the corresponding learning algorithm is given.
\subsection{TcGAN Framework}
In the TcGAN framework shown in Fig. \ref{Model}, the design of $ G_T $ is inspired by the long-range dependencies of the vision transformer to learn the overall layout of the training image and preserve the visual consistency of the whole image. Then, the local features of the image are extracted by $ G_C $ based on CNNs and extended to the generated image to accomplish the unconditional image generation task in a shorter time.

As shown in Fig. \ref{Model}(a) and Fig. \ref{Model}(d), the overall architecture of our proposed TcGAN consists of three parts: (i)$ N $ stages with $ N $ generators $ G = \{G_1, G_2, \cdot\cdot\cdot, G_N\}$ and 1 discriminator $D$. (ii) $ G_T $ with $ n $ transformer encoders. (iii)  $ G_C $ with one-layer upsampling and three-layer convolution. Their design idea are described as follows: (i) As shown in Fig. \ref{Model}(a), our generator contains multiple stages, where each stage of the generator (denoted as $ G_i $) consists of $ n $ global networks and $ i $ local networks for generating images of different sizes, where $ i \in \{1,2,\cdot\cdot\cdot,N\} $. The dashed gray arrow in Fig. \ref{Model}(a) indicates the model expansion and parallel training mechanism, i.e., the local network structure is added on top of the generator $ G_{i-1} $ structure to construct $ G_i $ structure, and the weight parameters of $ G_{i-1}$ are frozen to effectively reduce the parameters and shorten the training time during model training. (ii) As shown in Fig. \ref{Model}(b), $ G_T $ is a network of transformer cascade structures, which is mainly used to capture long-range dependencies between image patches to learn high-level semantic information about the global structure. (iii) As shown in Fig. \ref{Model}(c), the goal of the upsampling operation in $ G_C $ is to resize the generated image. Then a three-layer convolution operation is performed so as to model its local features and texture characteristics to enhance the fidelity of the output image. (iv) Based on the concept of whole before detail, we propose to perform the OSG task by firstly sampling the overall appearance to maintain the visual consistency between adjacent images, and then efficiently extract the local information to generate realistic images. Therefore, TcGAN performs $ G_T $ first and then trains $ G_C $, and both networks have the same objective function during training and follow the same overall appearance as the training images. The above process with input noise $ z $ can be described as follows:
\begin{equation}
	\label{equal0}
	{\bf{W}} = {G_T}(z)
\end{equation}
\begin{equation}
	\label{equal1}
	{\bf{fake}}_i = {G_C^i}({\bf{W}})
\end{equation}
where $ z \in \mathbb{R}^{S} , {\bf{W}} \in \mathbb{R}^{H \times L \times C} $, $ {\bf{fake}}_i \in \mathbb{R}^{H^{'} \times L^{'} \times C} $, $ S $ defaults to 1024, $ i $ denotes the $i$-th stage, $ G_C^i $ and $ {\bf{fake}}_i $ respectively denotes the local network and the generated image of the $i$-th stage.

\subsection{Scaling Formulation}
\begin{figure}[h]
	\centering
	\includegraphics[width=3.5in]{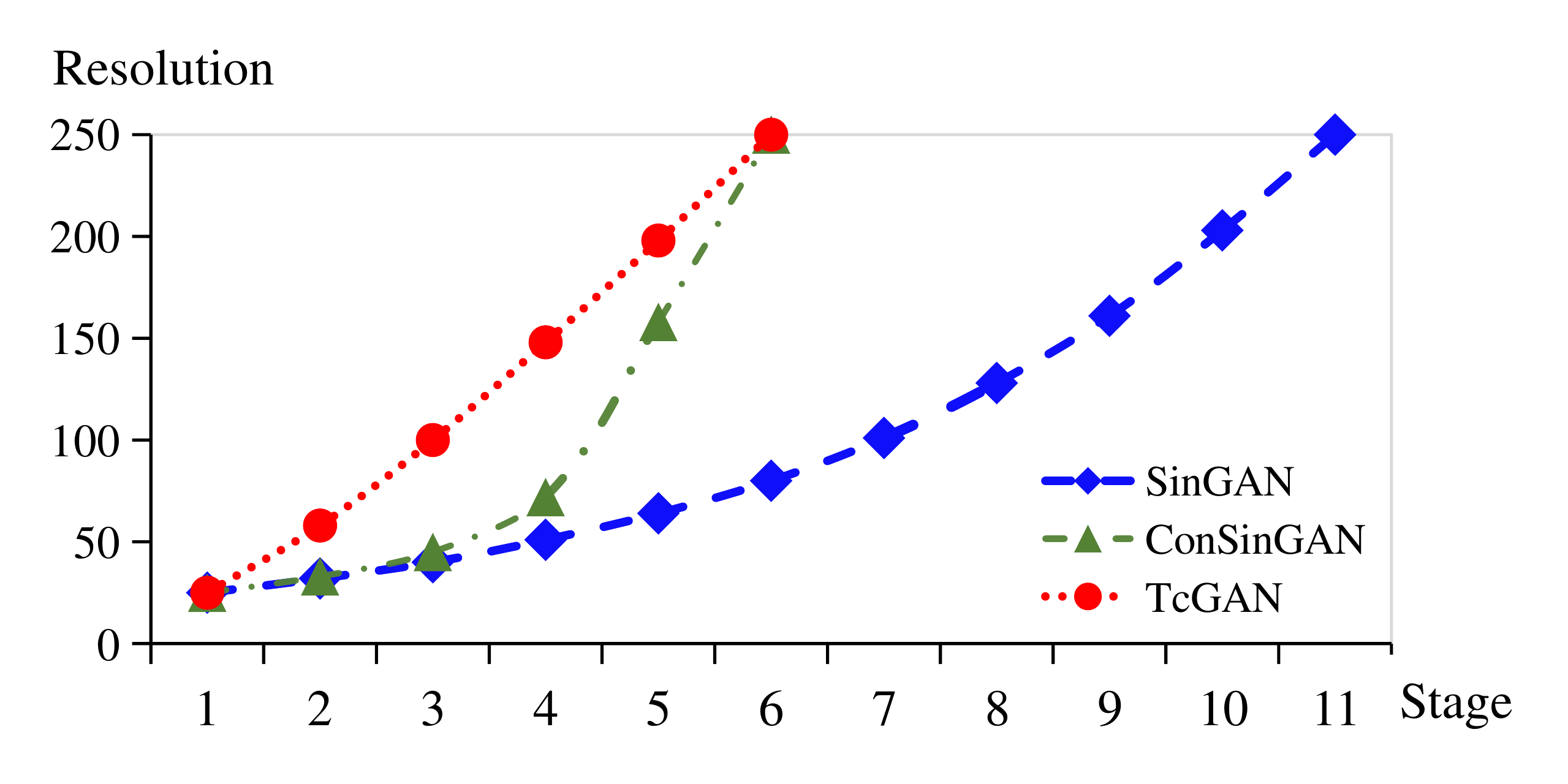}
	\caption{Comparison results of three scaling methods for outputting images with a maximum size of 250.}
	\label{fig14}
\end{figure}

SinGAN\cite{shaham2019singan9} proposed to downsample the training image $ {\bf{x}}$ using a scaling factor $r$. ConSinGAN\cite{hinz2021improved28} proposed to achieve the alleged global image layout by scaling as many low-resolution images $ {\bf{x}}$ as possible. However, it only works better in image generation with a resolution of 250, because on super-resolution tasks with increasing size, focusing too closely on the number of low-resolution images leads to a dramatic acceleration in the size of the super-resolution images generated in the high-stage, resulting in losing local feature information. At this point, this problem can be mitigated by increasing the number of stages by a certain amount, but leading to higher time complexity. Hence, based on the TcGAN architecture, we propose a novel scaling method for the basic OSG task while extending it to super-resolution tasks, which effectively improves the generalization capability of TcGAN. The reason for this design is that TcGAN displays high-level semantic-aware information that expresses global structure during image generation, which requires more attention to the effective extraction of local features as the resolution increases when applied to high-resolution tasks. The proposed scaling method is described as:
\begin{equation}
	\label{equal8}
	{\bf{x}}_i = {\bf{x}}_1 \times (1 + r \times (i+1) \times \frac{\ln{i}}{1+e^{-i}})
\end{equation}
where $ i \in \{1,2,\cdot\cdot\cdot,N\} $. As shown in Fig. \ref{fig14}, when rescaling scalar $r$=0.72, we observe that the new rescaling method can steadily increase the distance between the before and after resolutions compared with SinGAN and ConSinGAN.

\subsection{Global Network ($ G_T $)}
The goal of $G_T $ is to define a display representation for the global structure of the generated image in the OSG method to obtain its high-level semantic-aware information, rather than through a rough definition of the low-stage image leading to phenomena such as illusions and overlaps. Therefore, we propose a global network with a individual vision transformer to learn the global information representation of the training image, capturing the long-range dependencies between image patches, as shown in Fig. \ref{Model}(b). The original input of the standard transformer\cite{vaswani2017attention11} in the NLP task is a sentence, which is transformed into a sequence of one-dimensional token embeddings after a series of operations of padding, padding mask, and position encoding before being used as the input of the transformer encoder. In the image classification task, the original input of the vision transformer\cite{dosovitskiy2020image16} is an image, which is transformed into a patch embedding after slicing, encoding, and other operations before it is used as the input of the transformer encoder. However, in this image generation task, the original input is a one-dimensional Gaussian noise $ z $. Similarly, it is transformed into patches after a series of operations such as mapping and position embeddings before it is used as input to the transformer encoder. The first layer is the mapping operation which is composed of one-layer mapping network (denoted as $ \text{MLP}_{1} $), whose input is a Gaussian noise vector $ z $ and output is a potential vector {\bf{g}}, as shown in  Eq.\ref{equal3}:
\begin{equation}
	\label{equal3}
	{\bf{g}} = \text{MLP}_{1}(z)
\end{equation}
where $ {\bf{g}} \in \mathbb{R}^{min \times L}, z \in \mathbb{R}^{S} $. The next operation is the generated position embeddings sequence which is also used as input as shown in Eq.\ref{equal4}:
\begin{equation}
	\label{equal4}
	{\bf{h}}_{0} = {\bf{E}}_{pos}
\end{equation}
where $ {\bf{h}}_{0} \in \mathbb{R}^{min \times L} $. The first input received in the transformer architecture is normalized, where we use the self-model LayerNorm (SLN)\cite{lee2021vitgan24} usually used in previous studies. This benefits from the fact that the SLN is independent of external information and depends only on the input object, which is calculated as shown in Eq.\ref{equal2}:
\begin{equation}
	\label{equal2}
	\text{SLN}({\bf{h}}_0, {\bf{g}}) = {\alpha}_k({\bf{g}}) + {\beta}_k({\bf{g}}) \odot (\frac{{{\bf{h}}}_0 - {\boldsymbol{\mu}}}{\boldsymbol{\sigma}})
\end{equation}
where $ \boldsymbol{\mu} $ and $ \boldsymbol{\sigma} $ denote the mean and variance of the input layer, $ {\bf{\alpha}}_k $ and $ \beta_k $ denote the adaptive normalization parameters of the potential vector $ {\bf{g}} $, and $ \odot $ denotes the dot product. SLN is applied before every multihead self-attention (MSA) and MLP block, and residual connections after every block. The potential vector $ {\bf{g}} $ after SLN are split into $ N_h $ heads $ {\bf{g}}_{h} \in \mathbb{R}^{N_h \times min \times (\frac{L}{N_h})} $. In each head, given learnable matrices $ {\bf{W}}_q $, $ {\bf{W}}_k $, $ {\bf{W}}_v $, $ {\bf{g}}_{h^{'}} \in \mathbb{R}^{min \times (\frac{L}{N_h})}$ is used to calculated $ {\bf{Q}}_h $, $ {\bf{K}}_h $ and $ {\bf{V}}_h $ by linear layers, where $ h \in \{1,2,\cdot\cdot\cdot,N_h\}$ and $ N_h $ denotes the number of heads as the default of 6, as shown in Eqs.\ref{equal12A}-\ref{equal12c}:
\begin{subequations}\label{equal12}
	\begin{align}
		{\bf{Q}}_h &={\bf{g}}_{h^{'}}{\bf{W}}_q  \label{equal12A}\\
		{\bf{K}}_h &={\bf{g}}_{h^{'}}{\bf{W}}_k  \label{equal12B}\\
		{\bf{V}}_h &= {\bf{g}}_{h^{'}}{\bf{W}}_v \label{equal12c}
	\end{align}
\end{subequations}
then a single head self-attention can be calculated according to Eq.\ref{equal6}:
\begin{equation}
	\label{equal6}
	\text{Attention}_h({\bf{g}}_{h^{'}}) = \text{softmax}(\frac{{{\bf{Q}}_h{\bf{K}}}_h^T}{\sqrt{d_h}}){\bf{V}}_h
\end{equation}
where $ \sqrt{d_h} $ denotes the feature dimension of each head. The results of each head are then concatenated and linearly projected to obtain the multi-headed self-attention, as shown in Eq.\ref{equal13}:
\begin{equation}
	\label{equal13}
	\text{MSA}({\bf{g}}_{h^{'}}) = \text{concat}_{h=1}^{N_h}[\text{Attention}_h({\bf{g}}_{h^{'}})]{\bf{W}} + {\bf{b}}
\end{equation}
then the residual connections are applied to the above process. Finally, a display representation of the global information is obtained after the MLP layer. All the processes of the above $ G_T $ is obtained by Eq.\ref{equal5}:
\begin{equation}
	\begin{aligned}
		\label{equal5}
		{\bf{h}}_{k}& = {G_T(z)} = {\bf{h}}_{k-1} + \text{Drop}(\text{MSA}(\text{SLN}({\bf{h}}_{k-1}, {\bf{g}}))) + \\
		&\text{Drop}(\text{MLP}(\text{SLN}({\bf{h}}_{k-1} + \text{Drop}(\text{MSA}(\text{SLN}({\bf{h}}_{k-1}, {\bf{g}}))), {\bf{g}})))
	\end{aligned}
\end{equation}
where $ k \in \{1,2,\cdot\cdot\cdot,n\} $,  $ n $ denotes the number of $ G_T $, Drop denotes Dropout strategy which is often used to alleviate the overfitting problem.
\subsection{Local Network ($ G_C $)}
$ G_C$ consists of $\{ G_C^1, G_C^2, \cdot\cdot\cdot, G_C^N\}$, where the goal of $ G_C^i$ is to resize the output image with upsampling, while modeling the local features and texture information of the global structure results generated by $ G_T$ to improve the high fidelity of the output image (denoted as $ {\bf{fake}}_i $). In addition, the input of the $ G_C $ includes noisy images (denoted as {\bf{Z}} ) generated by Gaussian distribution to capture the fine-grained variations of the training images to increase the diversity of $ { \bf{fake}}_i $. The training process can be calculated from Eq.\ref{equal7} as follows:
\begin{equation}
	\label{equal7}
	{\bf{fake}}_i = G_C^i({\bf{h}}_{k^{'}}, {\bf{Z}})
\end{equation}
where $ {\bf{fake}}_i $ denotes the fake image generated at the $ i $-th stage, $ {\bf{h}}_{k^{'}} $ denotes the recombination form of $ {\bf{h}}_{k} $, $ {\bf{h}}_{k}\in \mathbb{R}^{min \times L} $, and $ {\bf{h}}_{{k}^{'}}\in \mathbb{R}^{min \times min \times C } $.
\begin{algorithm}		
	\caption{Overall TcGAN Process. \\  The Parameters of the Implemented Method in the Paper are Defaulted as: $iter$=1000, $N$=6, $n_{deep}$=3}\label{alg:alg1}
	\begin{algorithmic}[1]
		\State \textit{Require:} $N$ number of stages, $iter$ number of iterations, $n_{deep}$ deepth of training, $W_G^{i}$ the $i$-th Generator weights,  $W_D$ the Discriminator weights, $\phi_D$ the parameter Discriminator, $\phi_G^{i}$ the parameter of the $i$-th Generator.
		\State Sample random image {\bf{x}}.
		\State According to the proposed scaling formula,  {\bf{x}} is scaled into ${\bf{x}}_{i}=\{{{\bf{x}}_{1}, {\bf{x}}_2, {\bf{x}}_3, {\bf{x}}_4, {\bf{x}}_5, {\bf{x}}_6}\}$.
		\State Initialise $\phi_G^{1}$ and $\phi_D$.
		\State {\bf{for}} $i$ \textless \space  $ N $ {\bf{do}}
		\State \hspace{0.4cm} {\bf{for}} $ j $ \textless \space $iter$ {\bf{do}}
		\State \hspace{0.8cm} \# Training Generator $G_{i}$
		\State \hspace{0.8cm} {\bf{for}} $ k $ \textless \space $ n_{deep}$ {\bf{do}}
		\State \hspace{1.2cm} Make random noise $z$.
		\State \hspace{1.2cm} Generate fake image ${\bf{fake}}_{i}$.
		\State \hspace{1.2cm} Perform step of Adam update on $W_{G}^{i}$ with gene-
		\Statex\hspace{1.2cm} rator loss between ${\bf{x}}_{i}$ and ${\bf{fake}}_i$.
		\State \hspace{0.8cm} {\bf{end for}}	
		\State \hspace{0.8cm} \# Training Discriminator $D$
		\State \hspace{0.8cm} Make random noise $z$.
		\State \hspace{0.8cm} Generate fake image ${\bf{fake}}_{i}$.
		\State \hspace{0.8cm} Perform step of Adam update on $W_{D}$ with discrim-
		\Statex \hspace{0.8cm} inator loss between ${\bf{x}}_{i}$ and ${\bf{fake}}_i$.
		\State \hspace{0.4cm} {\bf{end for}}
		\State \hspace{0.4cm} Copy the result of the $\phi_G^{i}$ argument to $\phi_G^{i+1}$.
		\State {\bf{end for}}
	\end{algorithmic}
	\label{alg1}
\end{algorithm}
\subsection{Improved Model Training}
Inspired by PatchGAN, we uses it's architecture as our discriminator, which keeps its structure constant during the training process and continues training based on the weight parameters from the previous stage structure, as shown in Fig. \ref{Model}(d). So $ D_i $ indicates that the discriminator in the $ i $-th stage has the same structure with different parameters from the other stages. The input of $ D_i$  is the generated image $ {\bf{fake}}_i $ or the scaled training image $ {\bf{x}}_i $ at the $i$-th stage, and the output is whether the patch in the input image are true or not. During training the generator $ G_i $ of the current stage freezes the structure and parameters of the generator $ G_{i-1} $ of the last stage. The adversarial loss function for TcGAN optimization is Eq.\ref{equal9}:
\begin{equation}
	\label{equal9} 
	\mathop{\min}_{G_i} \mathop{\max}_{D_i} L_{adv}(G_i, D_i) + \gamma L_{rec}( G_i)
\end{equation}
where $ i \in \{1,2,\cdot\cdot\cdot,N\}, \gamma $ is the weight, and the default value is 10. The WGAN-GP\cite{gulrajani2017improved47} which improves the stability of training is used to adversarial loss item $ L_{adv}(G_i, D_i) $. 
The reconstruction loss term $ L_{rec}( G_i) $ is mainly used to optimize the generator and reduce the loss between the generated image $ G_i(z) $ and the training image $ {\bf{x}}_i$, as shown in Eq.\ref{equal10}:
\begin{equation}
	\label{equal10} 
	L_{rec}( G_i) = \| G_i(z) - {\bf{x}}_i \|^2
\end{equation}
where $ i \in \{1,2,\cdot\cdot\cdot,N\} $, $ {\bf{x}}_i $ is a downsampled image of ${\bf{ x}}$ by the proposed scaling formula in Eq.\ref{equal8}.\\

The TcGAN training algorithm is given and the pseudocode is described in Algorithm 1.

\section{Experiments}
In this section, we evaluate the generalization capability of the proposed method TcGAN in terms of qualitatively and quantitatively on a variety of scene images including nature, artistic, and complicated landscapes as well as object images.

\begin{figure*}[t]
	\centering
	\includegraphics[width=7.16in]{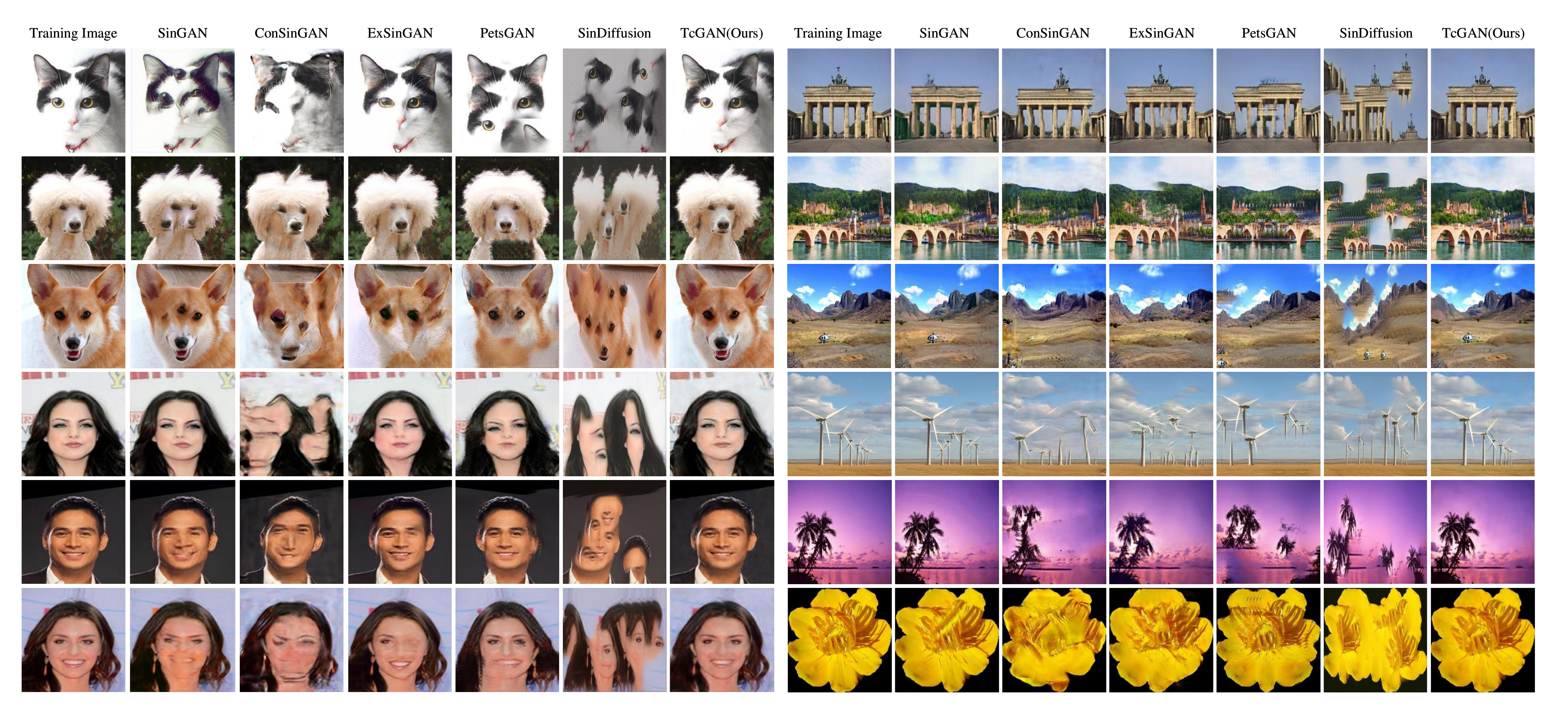}
	\caption{{\bf{Qualitative comparison.}} The visualization results of different OSG methods on three datasets. Compared with other methods, our method TcGAN not only achieves high-quality results on natural images but also achieves high-level semantic information consistency of global structure on object images. }
	\label{fig1}
\end{figure*}
\begin{figure*}[!t]
	\centering
	\includegraphics[width=7.16in]{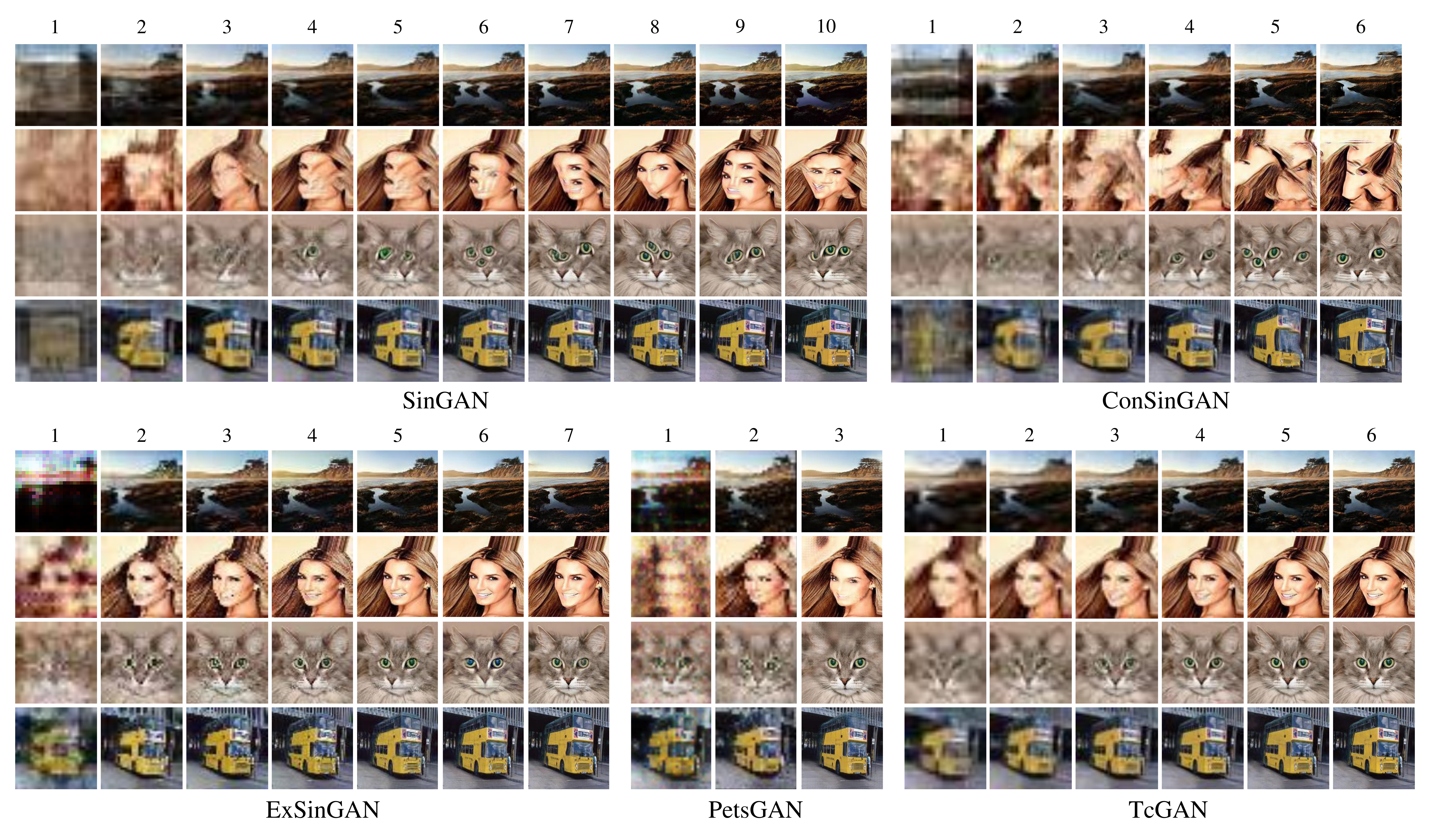}
	\caption{For different OSG methods of SinGAN, ConSinGAN, PetsGAN, ExSinGAN and TcGAN, the process of global structure and local details change in the implementation of the method in image generation is studied according to the output results of each stage. Images are randomly selected.}
	\label{fig5}
\end{figure*} 
\begin{figure}[!t]
	\centering
	\includegraphics[width=3.5in]{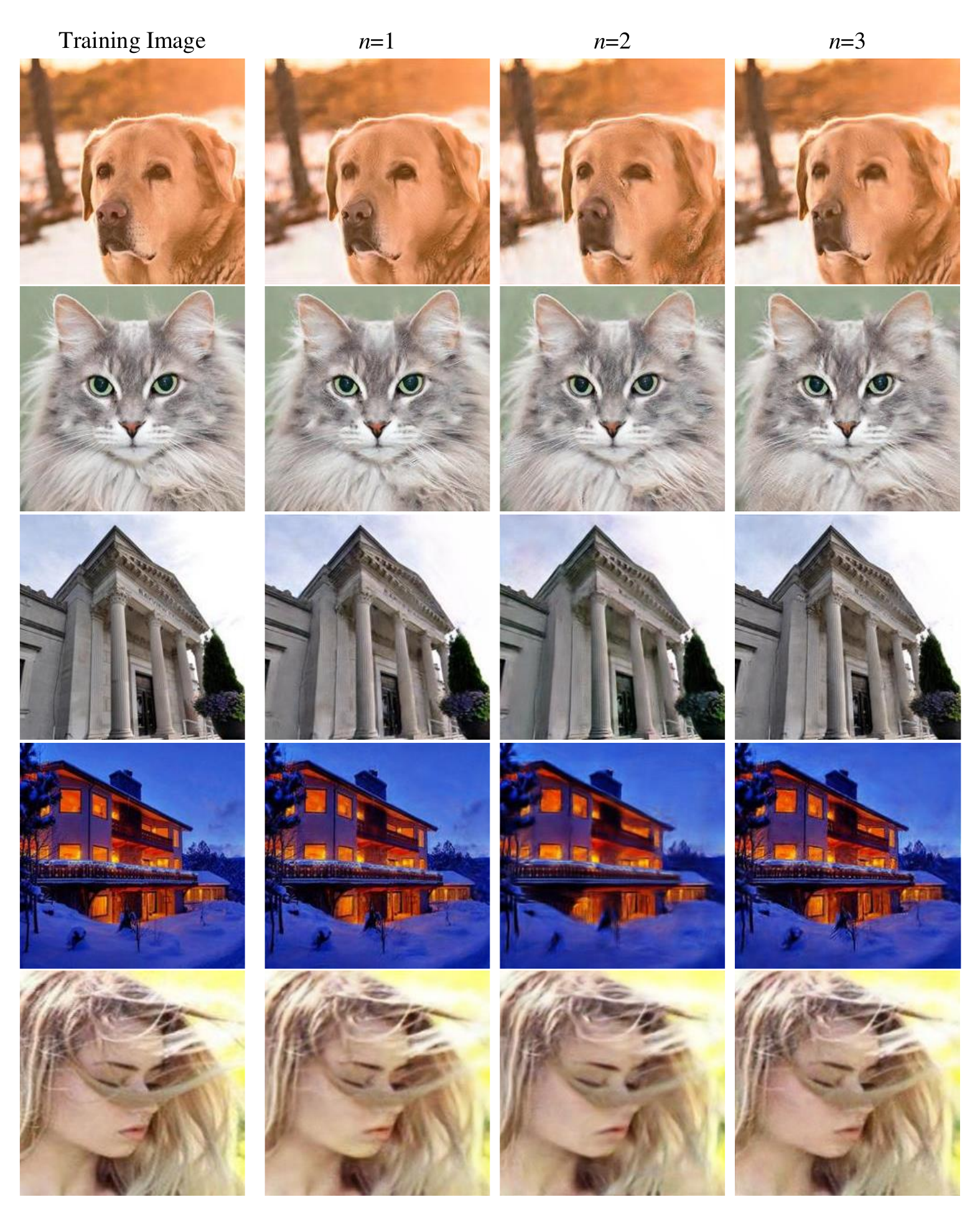}
	\caption{Study the effect of training with different number of $ G_T $ on TcGAN.}
	\label{fig4}
\end{figure} 
\begin{figure}[!t]
	\centering
	\includegraphics[width=3.5in]{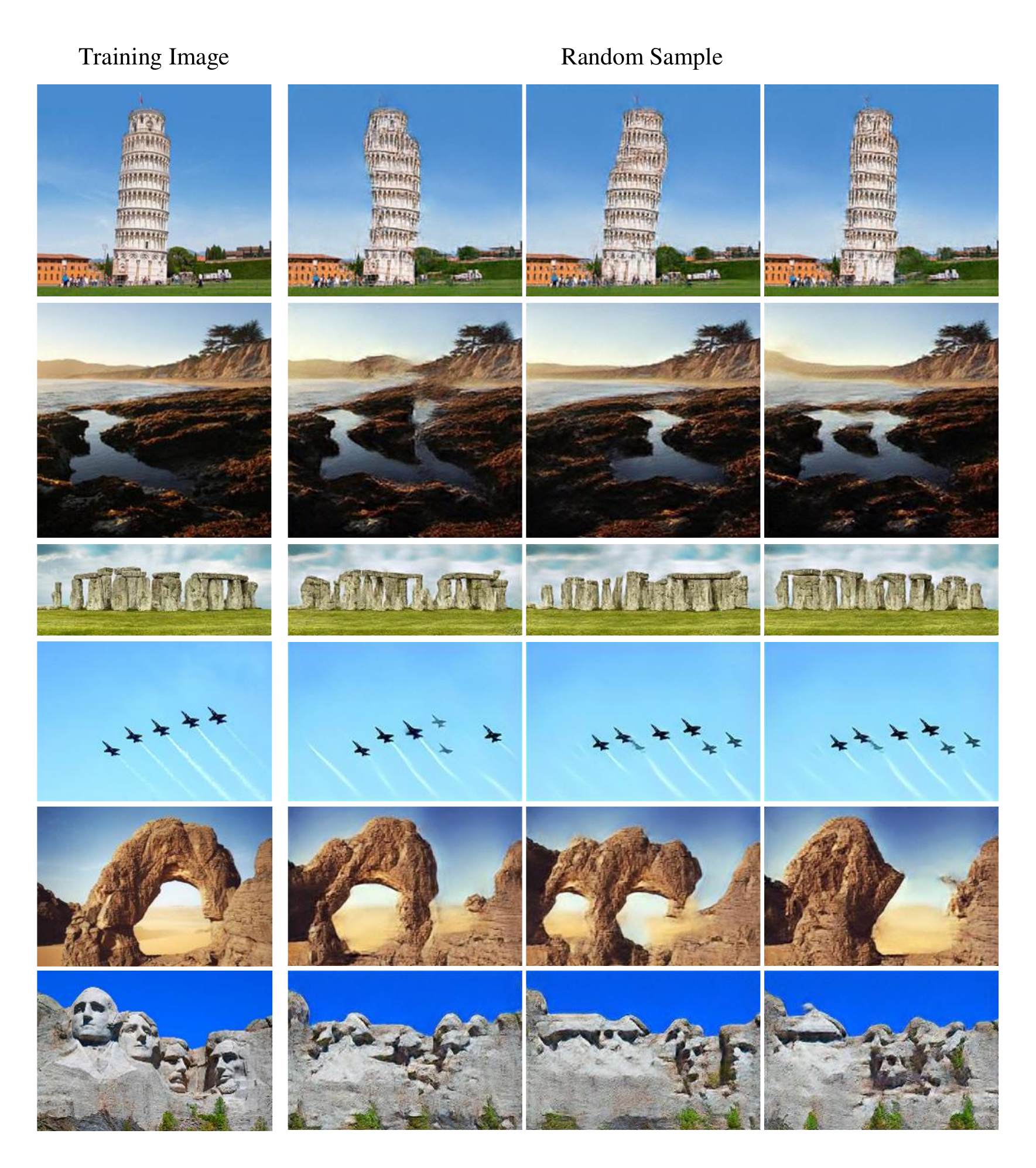}
	\caption{{\bf{Random Sample.}} Visualize the results of sample random experiments of TcGAN on the different datasets. }
	\label{fig10}
\end{figure}
\begin{figure}[!t]
	\centering
	\includegraphics[width=3.5in]{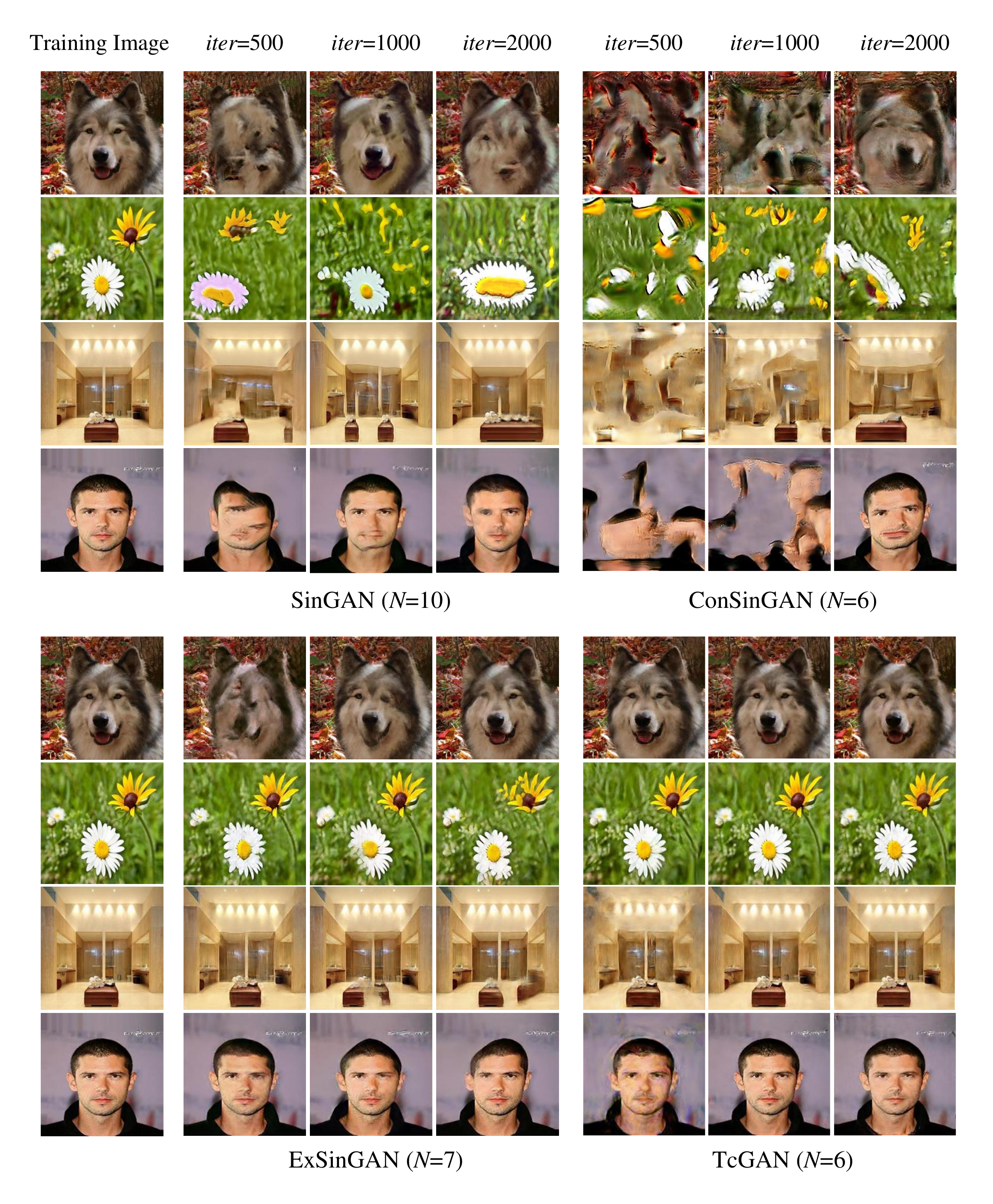}
	\caption{Compare the effects of SinGAN, ConSinGAN, ExSinGAN, and TcGAN on image generation under different number of $ iter $.}
	\label{fig3}
\end{figure}
\begin{figure}[!t]
	\centering
	\includegraphics[width=3.5in]{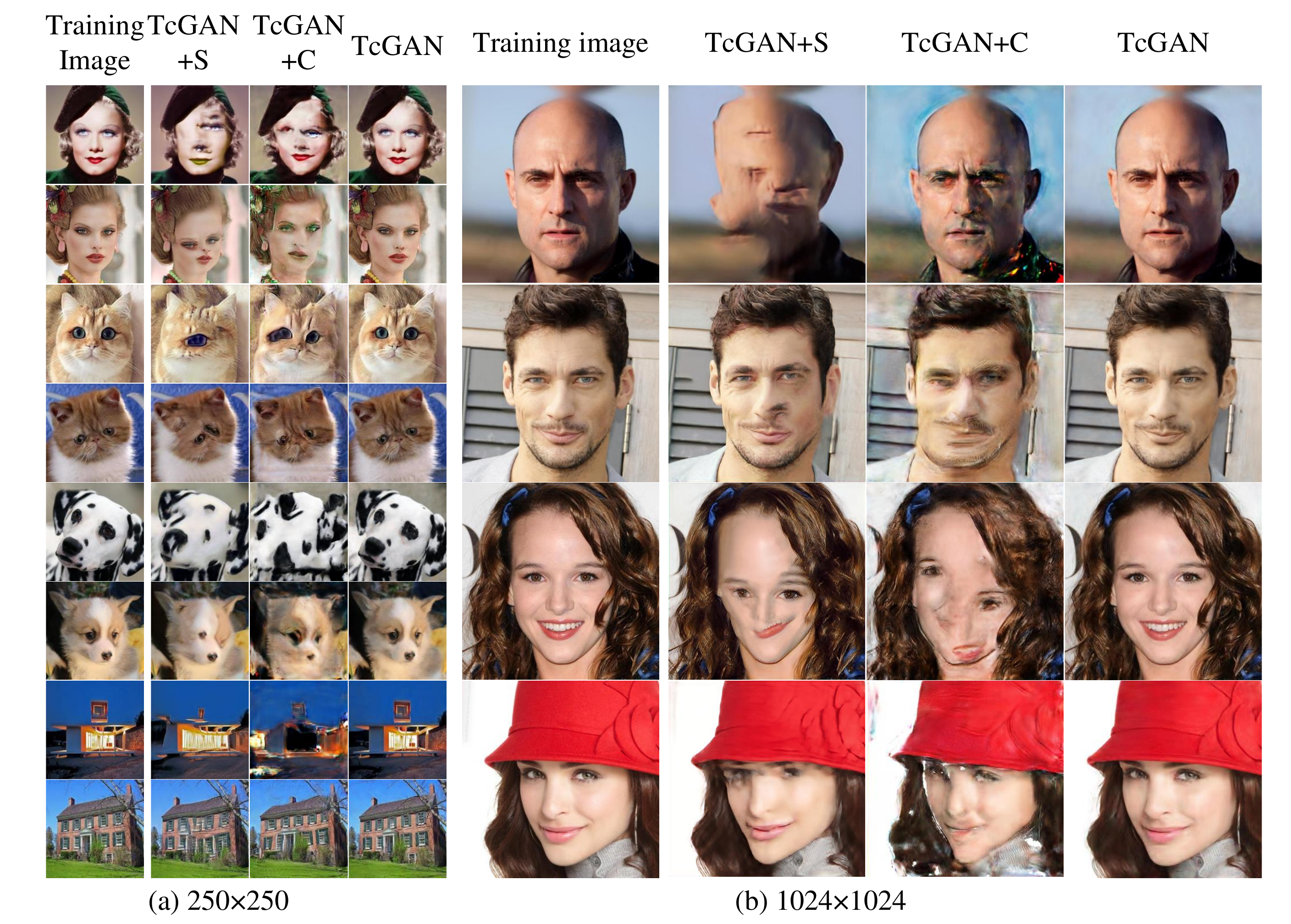}
	\caption{Compare the effects of TcGAN with different scaling formuilation.}
	\label{fig13}
\end{figure}
\begin{figure*}[!t]
	\centering
	\includegraphics[width=7.16in]{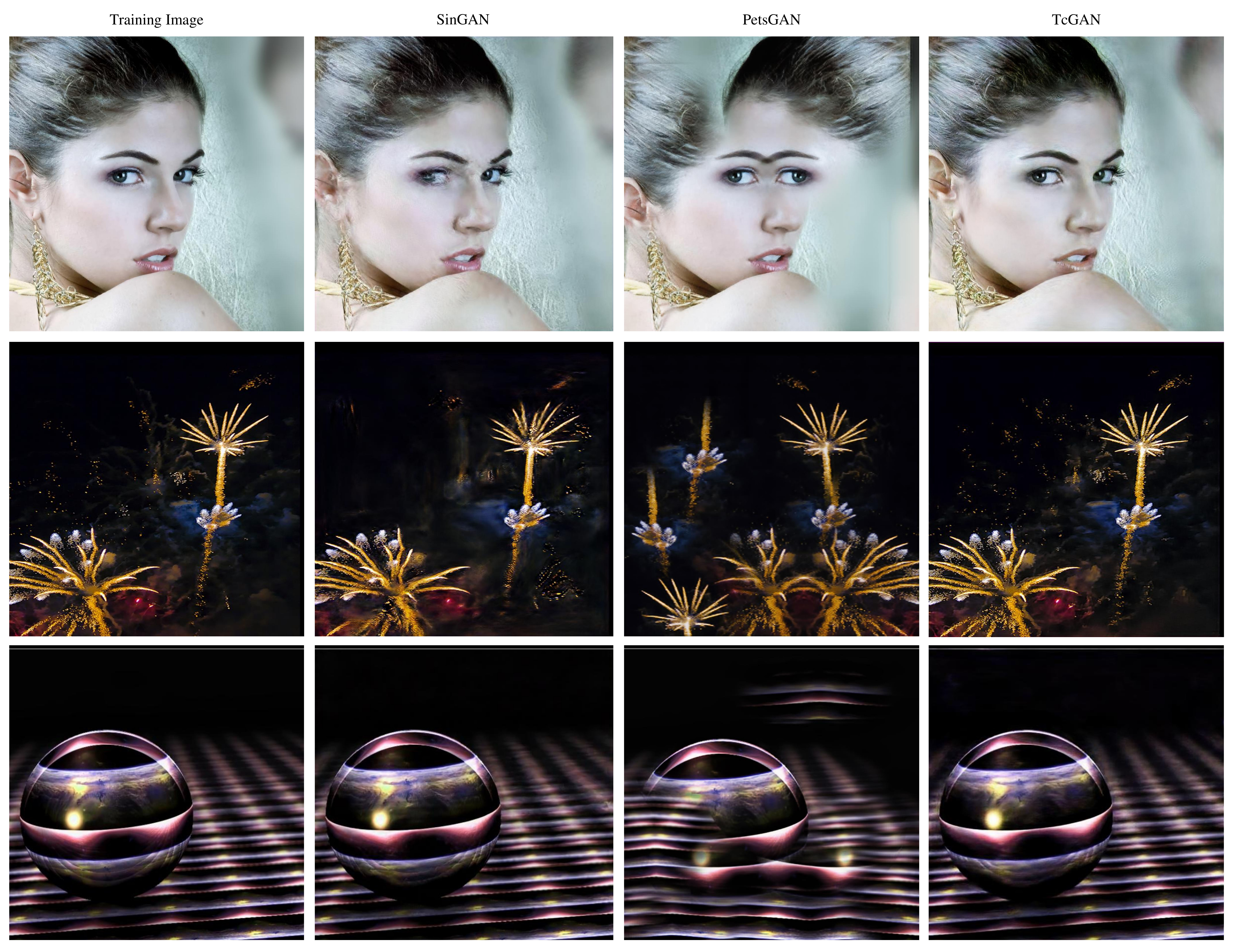}
	\caption{{\bf{Super-Resolution image generation.}} TcGAN trained on 1024×1024 images (top of the first column), 800×800 images (middle of the first column), and 600×600 images (bottom of the first column) realistically generates random samples with high-level semantic information and texture feature details of global structure.}
	\label{fig7}
\end{figure*}
\begin{figure}[!t]
	\centering
	\includegraphics[width=3.4in]{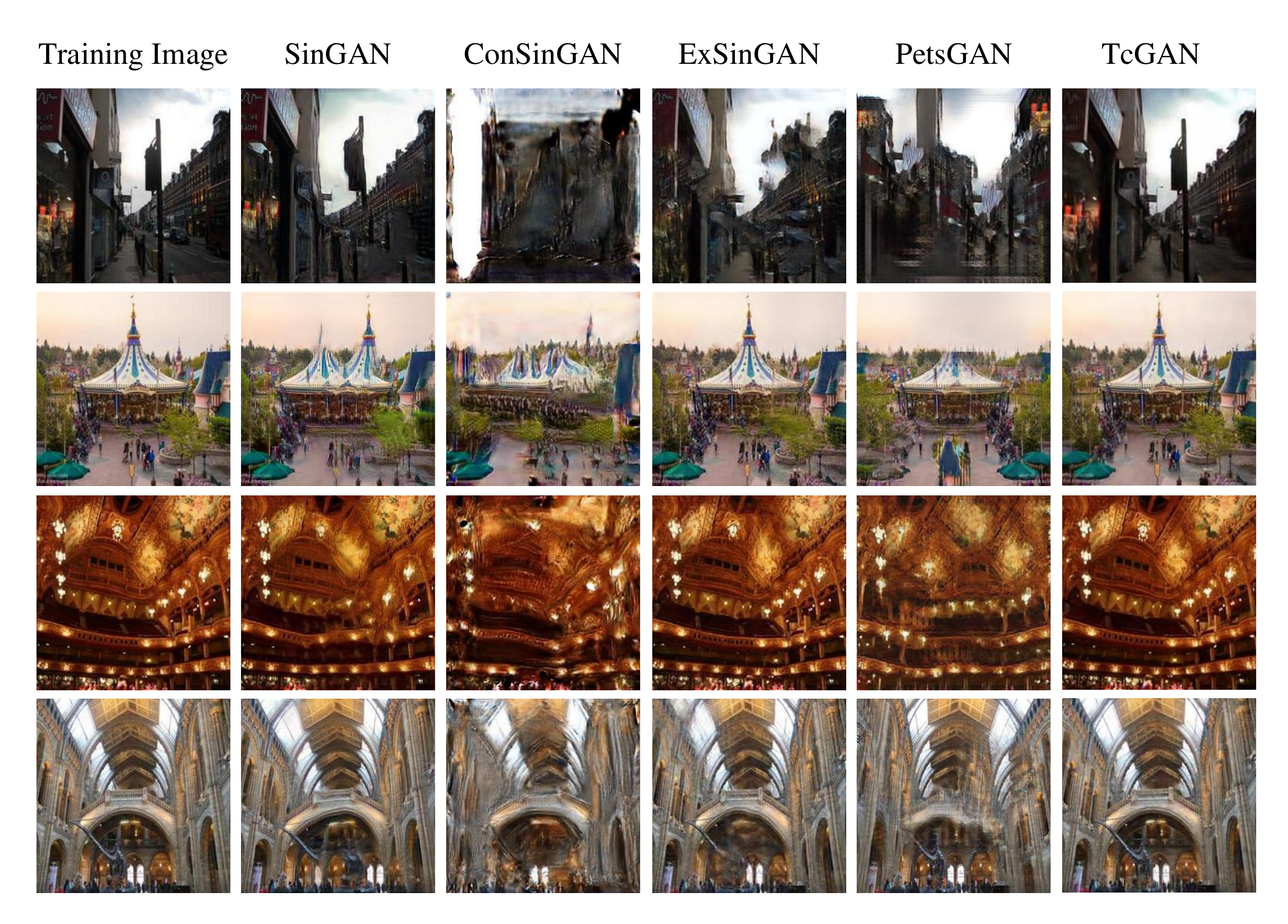}
	\caption{{\bf{Complicated Background.}} Visualize the results of comparative experiments of different OSG methods for complex global structures on the \emph{AFHQ50} dataset. }
	\label{fig2}
\end{figure}

\subsection{Implementations Setup}
\noindent {\bf{Datasets.}} \quad The images used in our experiments are \emph{AFHQ50}, \emph{Places50}, and \emph{CelebA50}, in which containing 50 images of animal faces, natural scenes and objects from the \emph{AFHQ}\cite{choi2020stargan48}, \emph{Places365}\cite{zhou2017places49}, and  \emph{CelebA}\cite{liu2015deep50} datasets, respectively. In addition, \emph{Flickr1024}\cite{Flickr102451}, \emph{DIV2K}\cite{DIV2K} and \emph{CelebA-HQ}\cite{karras2017progressive52} are used for super-resolution tasks.

\noindent {\bf{Baselines.}} \quad We evaluate our method on those datasets and compare the results with other state-of-the-art OSG methods, i.e., SinGAN\cite{shaham2019singan9}, ConSinGAN\cite{hinz2021improved28}, ExSinGAN\cite{zhang2021exsingan31}, PetsGAN\cite{zhang2022petsgan35}, TransGAN\cite{jiang2021transgan25}, SinDiffusion\cite{39_1}, and ViTGAN\cite{lee2021vitgan24} on image generation task, image synthesis, harmonization and super-resolution tasks. SinGAN, ConSinGAN, ExSinGAN, and PetsGAN are typical pure CNNs methods based on single image. However, TransGAN and ViTGAN are pure transformer methods. The training parameters of each of these methods are set as shown in Table \ref{tab1}:
\begin{table}[hb]
	\begin{center}
		\caption{Training parameters setting of the adopted methods}
		\label{tab1}
		\setlength{\tabcolsep}{0.7mm}{
			\begin{tabular}{|c|c|c|c|c|c|c|c|}
				\hline
				\bf{Methods} & \bf{Conv} & \bf{Pool} & \bf{Transformer} & $ {N}$ &  $ {iter} $ & $ {max} $ & Parameters\\
				\hline
				SinGAN & \resizebox{!}{\fontcharht\font`M}{\begin{tikzpicture}[y=0.80pt, x=0.80pt, yscale=-1.000000, xscale=1.000000, inner sep=0pt, outer sep=0pt]
\begin{scope}[shift={(100.0,2042.0)},nonzero rule]
  \path[draw=.,fill=.,line width=1.600pt] (1494.0000,-1516.0000) ..
    controls (1303.3333,-1381.3333) and (1125.6667,-1211.3333) ..
    (961.0000,-1006.0000) .. controls (790.3333,-792.6667) and
    (666.6667,-581.0000) .. (590.0000,-371.0000) -- (546.0000,-342.0000) ..
    controls (506.6667,-316.0000) and (474.0000,-291.3333) .. (448.0000,-268.0000)
    .. controls (443.3333,-292.6667) and (430.3333,-331.3333) ..
    (409.0000,-384.0000) -- (386.0000,-441.0000) .. controls (347.3333,-537.0000)
    and (318.0000,-603.0000) .. (298.0000,-639.0000) .. controls
    (248.0000,-727.0000) and (198.0000,-773.6667) .. (148.0000,-779.0000) ..
    controls (202.0000,-828.3333) and (249.0000,-853.0000) .. (289.0000,-853.0000)
    .. controls (343.6667,-853.0000) and (404.3333,-778.6667) ..
    (471.0000,-630.0000) -- (507.0000,-550.0000) .. controls (755.6667,-996.0000)
    and (1074.6667,-1335.3333) .. (1464.0000,-1568.0000) -- (1494.0000,-1516.0000)
    -- cycle;
\end{scope}

\end{tikzpicture}} & \resizebox{!}{\fontcharht\font`M}{\begin{tikzpicture}[y=0.80pt, x=0.80pt, yscale=-1.000000, xscale=1.000000, inner sep=0pt, outer sep=0pt]
\begin{scope}[shift={(100.0,2042.0)},nonzero rule]
  \path[draw=.,fill=.,line width=1.600pt] (1494.0000,-1516.0000) ..
    controls (1303.3333,-1381.3333) and (1125.6667,-1211.3333) ..
    (961.0000,-1006.0000) .. controls (790.3333,-792.6667) and
    (666.6667,-581.0000) .. (590.0000,-371.0000) -- (546.0000,-342.0000) ..
    controls (506.6667,-316.0000) and (474.0000,-291.3333) .. (448.0000,-268.0000)
    .. controls (443.3333,-292.6667) and (430.3333,-331.3333) ..
    (409.0000,-384.0000) -- (386.0000,-441.0000) .. controls (347.3333,-537.0000)
    and (318.0000,-603.0000) .. (298.0000,-639.0000) .. controls
    (248.0000,-727.0000) and (198.0000,-773.6667) .. (148.0000,-779.0000) ..
    controls (202.0000,-828.3333) and (249.0000,-853.0000) .. (289.0000,-853.0000)
    .. controls (343.6667,-853.0000) and (404.3333,-778.6667) ..
    (471.0000,-630.0000) -- (507.0000,-550.0000) .. controls (755.6667,-996.0000)
    and (1074.6667,-1335.3333) .. (1464.0000,-1568.0000) -- (1494.0000,-1516.0000)
    -- cycle;
\end{scope}

\end{tikzpicture}} & \resizebox{!}{\fontcharht\font`M}{\begin{tikzpicture}[y=0.80pt, x=0.80pt, yscale=-1.000000, xscale=1.000000, inner sep=0pt, outer sep=0pt]
\begin{scope}[shift={(100.0,1731.0)},nonzero rule]
  \path[draw=.,fill=.,line width=1.600pt] (1271.0000,-1280.0000) ..
    controls (1271.0000,-1272.0000) and (1263.6667,-1261.0000) ..
    (1249.0000,-1247.0000) .. controls (1237.0000,-1235.6667) and
    (1213.6667,-1207.6667) .. (1179.0000,-1163.0000) -- (866.0000,-754.0000) --
    (929.0000,-627.0000) -- (1089.0000,-285.0000) .. controls
    (1083.0000,-271.0000) and (1073.6667,-261.3333) .. (1061.0000,-256.0000) --
    (1061.0000,-240.0000) .. controls (1060.3333,-239.3333) and
    (1052.3333,-234.3333) .. (1037.0000,-225.0000) .. controls
    (1025.6667,-217.6667) and (1020.0000,-211.6667) .. (1020.0000,-207.0000) ..
    controls (1020.0000,-201.0000) and (1023.3333,-191.3333) ..
    (1030.0000,-178.0000) .. controls (1028.0000,-166.0000) and
    (1019.3333,-154.3333) .. (1004.0000,-143.0000) .. controls
    (988.6667,-132.3333) and (974.6667,-127.0000) .. (962.0000,-127.0000) ..
    controls (948.6667,-127.0000) and (922.6667,-157.0000) .. (884.0000,-217.0000)
    -- (692.0000,-514.0000) -- (286.0000,115.0000) .. controls (264.6667,131.0000)
    and (229.3333,139.0000) .. (180.0000,139.0000) .. controls (168.6667,139.0000)
    and (154.3333,125.3333) .. (137.0000,98.0000) .. controls (116.3333,65.3333)
    and (106.0000,44.3333) .. (106.0000,35.0000) .. controls (106.0000,27.0000)
    and (108.0000,16.0000) .. (112.0000,2.0000) .. controls (118.0000,-18.6667)
    and (117.3333,-32.0000) .. (110.0000,-38.0000) .. controls (103.3333,-44.6667)
    and (100.0000,-49.6667) .. (100.0000,-53.0000) .. controls (100.0000,-64.3333)
    and (116.3333,-98.0000) .. (149.0000,-154.0000) -- (198.0000,-238.0000) ..
    controls (250.6667,-328.0000) and (367.6667,-482.6667) .. (549.0000,-702.0000)
    -- (573.0000,-731.0000) -- (422.0000,-1149.0000) .. controls
    (406.6667,-1191.6667) and (399.0000,-1215.0000) .. (399.0000,-1219.0000) ..
    controls (399.0000,-1225.0000) and (408.0000,-1239.3333) ..
    (426.0000,-1262.0000) -- (440.0000,-1274.0000) .. controls
    (444.6667,-1278.0000) and (449.3333,-1280.0000) .. (454.0000,-1280.0000) ..
    controls (461.3333,-1280.0000) and (472.3333,-1274.0000) ..
    (487.0000,-1262.0000) .. controls (495.0000,-1268.6667) and
    (506.0000,-1272.0000) .. (520.0000,-1272.0000) .. controls
    (524.0000,-1274.0000) and (530.0000,-1278.6667) .. (538.0000,-1286.0000) ..
    controls (544.6667,-1297.3333) and (552.3333,-1303.0000) ..
    (561.0000,-1303.0000) .. controls (573.6667,-1303.0000) and
    (586.0000,-1290.6667) .. (598.0000,-1266.0000) -- (753.0000,-958.0000) --
    (805.0000,-1022.0000) -- (905.0000,-1137.0000) .. controls
    (1029.6667,-1280.3333) and (1095.0000,-1352.0000) .. (1101.0000,-1352.0000) ..
    controls (1117.6667,-1352.0000) and (1141.6667,-1339.0000) ..
    (1173.0000,-1313.0000) -- (1196.0000,-1333.0000) .. controls
    (1205.3333,-1347.0000) and (1212.6667,-1354.0000) .. (1218.0000,-1354.0000) ..
    controls (1221.3333,-1354.0000) and (1225.3333,-1348.3333) ..
    (1230.0000,-1337.0000) .. controls (1232.6667,-1331.0000) and
    (1240.6667,-1320.6667) .. (1254.0000,-1306.0000) .. controls
    (1265.3333,-1294.0000) and (1271.0000,-1285.3333) .. (1271.0000,-1280.0000) --
    cycle;
\end{scope}

\end{tikzpicture}} & 11 & 2000 & 250×250 & 2.92M\\
				\hline
				ConSinGAN & \resizebox{!}{\fontcharht\font`M}{\begin{tikzpicture}[y=0.80pt, x=0.80pt, yscale=-1.000000, xscale=1.000000, inner sep=0pt, outer sep=0pt]
\begin{scope}[shift={(100.0,2042.0)},nonzero rule]
  \path[draw=.,fill=.,line width=1.600pt] (1494.0000,-1516.0000) ..
    controls (1303.3333,-1381.3333) and (1125.6667,-1211.3333) ..
    (961.0000,-1006.0000) .. controls (790.3333,-792.6667) and
    (666.6667,-581.0000) .. (590.0000,-371.0000) -- (546.0000,-342.0000) ..
    controls (506.6667,-316.0000) and (474.0000,-291.3333) .. (448.0000,-268.0000)
    .. controls (443.3333,-292.6667) and (430.3333,-331.3333) ..
    (409.0000,-384.0000) -- (386.0000,-441.0000) .. controls (347.3333,-537.0000)
    and (318.0000,-603.0000) .. (298.0000,-639.0000) .. controls
    (248.0000,-727.0000) and (198.0000,-773.6667) .. (148.0000,-779.0000) ..
    controls (202.0000,-828.3333) and (249.0000,-853.0000) .. (289.0000,-853.0000)
    .. controls (343.6667,-853.0000) and (404.3333,-778.6667) ..
    (471.0000,-630.0000) -- (507.0000,-550.0000) .. controls (755.6667,-996.0000)
    and (1074.6667,-1335.3333) .. (1464.0000,-1568.0000) -- (1494.0000,-1516.0000)
    -- cycle;
\end{scope}

\end{tikzpicture}} & \resizebox{!}{\fontcharht\font`M}{\begin{tikzpicture}[y=0.80pt, x=0.80pt, yscale=-1.000000, xscale=1.000000, inner sep=0pt, outer sep=0pt]
\begin{scope}[shift={(100.0,2042.0)},nonzero rule]
  \path[draw=.,fill=.,line width=1.600pt] (1494.0000,-1516.0000) ..
    controls (1303.3333,-1381.3333) and (1125.6667,-1211.3333) ..
    (961.0000,-1006.0000) .. controls (790.3333,-792.6667) and
    (666.6667,-581.0000) .. (590.0000,-371.0000) -- (546.0000,-342.0000) ..
    controls (506.6667,-316.0000) and (474.0000,-291.3333) .. (448.0000,-268.0000)
    .. controls (443.3333,-292.6667) and (430.3333,-331.3333) ..
    (409.0000,-384.0000) -- (386.0000,-441.0000) .. controls (347.3333,-537.0000)
    and (318.0000,-603.0000) .. (298.0000,-639.0000) .. controls
    (248.0000,-727.0000) and (198.0000,-773.6667) .. (148.0000,-779.0000) ..
    controls (202.0000,-828.3333) and (249.0000,-853.0000) .. (289.0000,-853.0000)
    .. controls (343.6667,-853.0000) and (404.3333,-778.6667) ..
    (471.0000,-630.0000) -- (507.0000,-550.0000) .. controls (755.6667,-996.0000)
    and (1074.6667,-1335.3333) .. (1464.0000,-1568.0000) -- (1494.0000,-1516.0000)
    -- cycle;
\end{scope}

\end{tikzpicture}} & \resizebox{!}{\fontcharht\font`M}{\begin{tikzpicture}[y=0.80pt, x=0.80pt, yscale=-1.000000, xscale=1.000000, inner sep=0pt, outer sep=0pt]
\begin{scope}[shift={(100.0,1731.0)},nonzero rule]
  \path[draw=.,fill=.,line width=1.600pt] (1271.0000,-1280.0000) ..
    controls (1271.0000,-1272.0000) and (1263.6667,-1261.0000) ..
    (1249.0000,-1247.0000) .. controls (1237.0000,-1235.6667) and
    (1213.6667,-1207.6667) .. (1179.0000,-1163.0000) -- (866.0000,-754.0000) --
    (929.0000,-627.0000) -- (1089.0000,-285.0000) .. controls
    (1083.0000,-271.0000) and (1073.6667,-261.3333) .. (1061.0000,-256.0000) --
    (1061.0000,-240.0000) .. controls (1060.3333,-239.3333) and
    (1052.3333,-234.3333) .. (1037.0000,-225.0000) .. controls
    (1025.6667,-217.6667) and (1020.0000,-211.6667) .. (1020.0000,-207.0000) ..
    controls (1020.0000,-201.0000) and (1023.3333,-191.3333) ..
    (1030.0000,-178.0000) .. controls (1028.0000,-166.0000) and
    (1019.3333,-154.3333) .. (1004.0000,-143.0000) .. controls
    (988.6667,-132.3333) and (974.6667,-127.0000) .. (962.0000,-127.0000) ..
    controls (948.6667,-127.0000) and (922.6667,-157.0000) .. (884.0000,-217.0000)
    -- (692.0000,-514.0000) -- (286.0000,115.0000) .. controls (264.6667,131.0000)
    and (229.3333,139.0000) .. (180.0000,139.0000) .. controls (168.6667,139.0000)
    and (154.3333,125.3333) .. (137.0000,98.0000) .. controls (116.3333,65.3333)
    and (106.0000,44.3333) .. (106.0000,35.0000) .. controls (106.0000,27.0000)
    and (108.0000,16.0000) .. (112.0000,2.0000) .. controls (118.0000,-18.6667)
    and (117.3333,-32.0000) .. (110.0000,-38.0000) .. controls (103.3333,-44.6667)
    and (100.0000,-49.6667) .. (100.0000,-53.0000) .. controls (100.0000,-64.3333)
    and (116.3333,-98.0000) .. (149.0000,-154.0000) -- (198.0000,-238.0000) ..
    controls (250.6667,-328.0000) and (367.6667,-482.6667) .. (549.0000,-702.0000)
    -- (573.0000,-731.0000) -- (422.0000,-1149.0000) .. controls
    (406.6667,-1191.6667) and (399.0000,-1215.0000) .. (399.0000,-1219.0000) ..
    controls (399.0000,-1225.0000) and (408.0000,-1239.3333) ..
    (426.0000,-1262.0000) -- (440.0000,-1274.0000) .. controls
    (444.6667,-1278.0000) and (449.3333,-1280.0000) .. (454.0000,-1280.0000) ..
    controls (461.3333,-1280.0000) and (472.3333,-1274.0000) ..
    (487.0000,-1262.0000) .. controls (495.0000,-1268.6667) and
    (506.0000,-1272.0000) .. (520.0000,-1272.0000) .. controls
    (524.0000,-1274.0000) and (530.0000,-1278.6667) .. (538.0000,-1286.0000) ..
    controls (544.6667,-1297.3333) and (552.3333,-1303.0000) ..
    (561.0000,-1303.0000) .. controls (573.6667,-1303.0000) and
    (586.0000,-1290.6667) .. (598.0000,-1266.0000) -- (753.0000,-958.0000) --
    (805.0000,-1022.0000) -- (905.0000,-1137.0000) .. controls
    (1029.6667,-1280.3333) and (1095.0000,-1352.0000) .. (1101.0000,-1352.0000) ..
    controls (1117.6667,-1352.0000) and (1141.6667,-1339.0000) ..
    (1173.0000,-1313.0000) -- (1196.0000,-1333.0000) .. controls
    (1205.3333,-1347.0000) and (1212.6667,-1354.0000) .. (1218.0000,-1354.0000) ..
    controls (1221.3333,-1354.0000) and (1225.3333,-1348.3333) ..
    (1230.0000,-1337.0000) .. controls (1232.6667,-1331.0000) and
    (1240.6667,-1320.6667) .. (1254.0000,-1306.0000) .. controls
    (1265.3333,-1294.0000) and (1271.0000,-1285.3333) .. (1271.0000,-1280.0000) --
    cycle;
\end{scope}

\end{tikzpicture}} & 6 & 2000 & 250×250	 & 0.78M \\
				\hline
				ExSinGAN & \resizebox{!}{\fontcharht\font`M}{\begin{tikzpicture}[y=0.80pt, x=0.80pt, yscale=-1.000000, xscale=1.000000, inner sep=0pt, outer sep=0pt]
\begin{scope}[shift={(100.0,2042.0)},nonzero rule]
  \path[draw=.,fill=.,line width=1.600pt] (1494.0000,-1516.0000) ..
    controls (1303.3333,-1381.3333) and (1125.6667,-1211.3333) ..
    (961.0000,-1006.0000) .. controls (790.3333,-792.6667) and
    (666.6667,-581.0000) .. (590.0000,-371.0000) -- (546.0000,-342.0000) ..
    controls (506.6667,-316.0000) and (474.0000,-291.3333) .. (448.0000,-268.0000)
    .. controls (443.3333,-292.6667) and (430.3333,-331.3333) ..
    (409.0000,-384.0000) -- (386.0000,-441.0000) .. controls (347.3333,-537.0000)
    and (318.0000,-603.0000) .. (298.0000,-639.0000) .. controls
    (248.0000,-727.0000) and (198.0000,-773.6667) .. (148.0000,-779.0000) ..
    controls (202.0000,-828.3333) and (249.0000,-853.0000) .. (289.0000,-853.0000)
    .. controls (343.6667,-853.0000) and (404.3333,-778.6667) ..
    (471.0000,-630.0000) -- (507.0000,-550.0000) .. controls (755.6667,-996.0000)
    and (1074.6667,-1335.3333) .. (1464.0000,-1568.0000) -- (1494.0000,-1516.0000)
    -- cycle;
\end{scope}

\end{tikzpicture}} & \resizebox{!}{\fontcharht\font`M}{\begin{tikzpicture}[y=0.80pt, x=0.80pt, yscale=-1.000000, xscale=1.000000, inner sep=0pt, outer sep=0pt]
\begin{scope}[shift={(100.0,2042.0)},nonzero rule]
  \path[draw=.,fill=.,line width=1.600pt] (1494.0000,-1516.0000) ..
    controls (1303.3333,-1381.3333) and (1125.6667,-1211.3333) ..
    (961.0000,-1006.0000) .. controls (790.3333,-792.6667) and
    (666.6667,-581.0000) .. (590.0000,-371.0000) -- (546.0000,-342.0000) ..
    controls (506.6667,-316.0000) and (474.0000,-291.3333) .. (448.0000,-268.0000)
    .. controls (443.3333,-292.6667) and (430.3333,-331.3333) ..
    (409.0000,-384.0000) -- (386.0000,-441.0000) .. controls (347.3333,-537.0000)
    and (318.0000,-603.0000) .. (298.0000,-639.0000) .. controls
    (248.0000,-727.0000) and (198.0000,-773.6667) .. (148.0000,-779.0000) ..
    controls (202.0000,-828.3333) and (249.0000,-853.0000) .. (289.0000,-853.0000)
    .. controls (343.6667,-853.0000) and (404.3333,-778.6667) ..
    (471.0000,-630.0000) -- (507.0000,-550.0000) .. controls (755.6667,-996.0000)
    and (1074.6667,-1335.3333) .. (1464.0000,-1568.0000) -- (1494.0000,-1516.0000)
    -- cycle;
\end{scope}

\end{tikzpicture}} & \resizebox{!}{\fontcharht\font`M}{\begin{tikzpicture}[y=0.80pt, x=0.80pt, yscale=-1.000000, xscale=1.000000, inner sep=0pt, outer sep=0pt]
\begin{scope}[shift={(100.0,1731.0)},nonzero rule]
  \path[draw=.,fill=.,line width=1.600pt] (1271.0000,-1280.0000) ..
    controls (1271.0000,-1272.0000) and (1263.6667,-1261.0000) ..
    (1249.0000,-1247.0000) .. controls (1237.0000,-1235.6667) and
    (1213.6667,-1207.6667) .. (1179.0000,-1163.0000) -- (866.0000,-754.0000) --
    (929.0000,-627.0000) -- (1089.0000,-285.0000) .. controls
    (1083.0000,-271.0000) and (1073.6667,-261.3333) .. (1061.0000,-256.0000) --
    (1061.0000,-240.0000) .. controls (1060.3333,-239.3333) and
    (1052.3333,-234.3333) .. (1037.0000,-225.0000) .. controls
    (1025.6667,-217.6667) and (1020.0000,-211.6667) .. (1020.0000,-207.0000) ..
    controls (1020.0000,-201.0000) and (1023.3333,-191.3333) ..
    (1030.0000,-178.0000) .. controls (1028.0000,-166.0000) and
    (1019.3333,-154.3333) .. (1004.0000,-143.0000) .. controls
    (988.6667,-132.3333) and (974.6667,-127.0000) .. (962.0000,-127.0000) ..
    controls (948.6667,-127.0000) and (922.6667,-157.0000) .. (884.0000,-217.0000)
    -- (692.0000,-514.0000) -- (286.0000,115.0000) .. controls (264.6667,131.0000)
    and (229.3333,139.0000) .. (180.0000,139.0000) .. controls (168.6667,139.0000)
    and (154.3333,125.3333) .. (137.0000,98.0000) .. controls (116.3333,65.3333)
    and (106.0000,44.3333) .. (106.0000,35.0000) .. controls (106.0000,27.0000)
    and (108.0000,16.0000) .. (112.0000,2.0000) .. controls (118.0000,-18.6667)
    and (117.3333,-32.0000) .. (110.0000,-38.0000) .. controls (103.3333,-44.6667)
    and (100.0000,-49.6667) .. (100.0000,-53.0000) .. controls (100.0000,-64.3333)
    and (116.3333,-98.0000) .. (149.0000,-154.0000) -- (198.0000,-238.0000) ..
    controls (250.6667,-328.0000) and (367.6667,-482.6667) .. (549.0000,-702.0000)
    -- (573.0000,-731.0000) -- (422.0000,-1149.0000) .. controls
    (406.6667,-1191.6667) and (399.0000,-1215.0000) .. (399.0000,-1219.0000) ..
    controls (399.0000,-1225.0000) and (408.0000,-1239.3333) ..
    (426.0000,-1262.0000) -- (440.0000,-1274.0000) .. controls
    (444.6667,-1278.0000) and (449.3333,-1280.0000) .. (454.0000,-1280.0000) ..
    controls (461.3333,-1280.0000) and (472.3333,-1274.0000) ..
    (487.0000,-1262.0000) .. controls (495.0000,-1268.6667) and
    (506.0000,-1272.0000) .. (520.0000,-1272.0000) .. controls
    (524.0000,-1274.0000) and (530.0000,-1278.6667) .. (538.0000,-1286.0000) ..
    controls (544.6667,-1297.3333) and (552.3333,-1303.0000) ..
    (561.0000,-1303.0000) .. controls (573.6667,-1303.0000) and
    (586.0000,-1290.6667) .. (598.0000,-1266.0000) -- (753.0000,-958.0000) --
    (805.0000,-1022.0000) -- (905.0000,-1137.0000) .. controls
    (1029.6667,-1280.3333) and (1095.0000,-1352.0000) .. (1101.0000,-1352.0000) ..
    controls (1117.6667,-1352.0000) and (1141.6667,-1339.0000) ..
    (1173.0000,-1313.0000) -- (1196.0000,-1333.0000) .. controls
    (1205.3333,-1347.0000) and (1212.6667,-1354.0000) .. (1218.0000,-1354.0000) ..
    controls (1221.3333,-1354.0000) and (1225.3333,-1348.3333) ..
    (1230.0000,-1337.0000) .. controls (1232.6667,-1331.0000) and
    (1240.6667,-1320.6667) .. (1254.0000,-1306.0000) .. controls
    (1265.3333,-1294.0000) and (1271.0000,-1285.3333) .. (1271.0000,-1280.0000) --
    cycle;
\end{scope}

\end{tikzpicture}} & 7 & 2000 & 250×250 & 21.35M \\
				\hline
				PetsGAN & \resizebox{!}{\fontcharht\font`M}{\begin{tikzpicture}[y=0.80pt, x=0.80pt, yscale=-1.000000, xscale=1.000000, inner sep=0pt, outer sep=0pt]
\begin{scope}[shift={(100.0,2042.0)},nonzero rule]
  \path[draw=.,fill=.,line width=1.600pt] (1494.0000,-1516.0000) ..
    controls (1303.3333,-1381.3333) and (1125.6667,-1211.3333) ..
    (961.0000,-1006.0000) .. controls (790.3333,-792.6667) and
    (666.6667,-581.0000) .. (590.0000,-371.0000) -- (546.0000,-342.0000) ..
    controls (506.6667,-316.0000) and (474.0000,-291.3333) .. (448.0000,-268.0000)
    .. controls (443.3333,-292.6667) and (430.3333,-331.3333) ..
    (409.0000,-384.0000) -- (386.0000,-441.0000) .. controls (347.3333,-537.0000)
    and (318.0000,-603.0000) .. (298.0000,-639.0000) .. controls
    (248.0000,-727.0000) and (198.0000,-773.6667) .. (148.0000,-779.0000) ..
    controls (202.0000,-828.3333) and (249.0000,-853.0000) .. (289.0000,-853.0000)
    .. controls (343.6667,-853.0000) and (404.3333,-778.6667) ..
    (471.0000,-630.0000) -- (507.0000,-550.0000) .. controls (755.6667,-996.0000)
    and (1074.6667,-1335.3333) .. (1464.0000,-1568.0000) -- (1494.0000,-1516.0000)
    -- cycle;
\end{scope}

\end{tikzpicture}} & \resizebox{!}{\fontcharht\font`M}{\begin{tikzpicture}[y=0.80pt, x=0.80pt, yscale=-1.000000, xscale=1.000000, inner sep=0pt, outer sep=0pt]
\begin{scope}[shift={(100.0,2042.0)},nonzero rule]
  \path[draw=.,fill=.,line width=1.600pt] (1494.0000,-1516.0000) ..
    controls (1303.3333,-1381.3333) and (1125.6667,-1211.3333) ..
    (961.0000,-1006.0000) .. controls (790.3333,-792.6667) and
    (666.6667,-581.0000) .. (590.0000,-371.0000) -- (546.0000,-342.0000) ..
    controls (506.6667,-316.0000) and (474.0000,-291.3333) .. (448.0000,-268.0000)
    .. controls (443.3333,-292.6667) and (430.3333,-331.3333) ..
    (409.0000,-384.0000) -- (386.0000,-441.0000) .. controls (347.3333,-537.0000)
    and (318.0000,-603.0000) .. (298.0000,-639.0000) .. controls
    (248.0000,-727.0000) and (198.0000,-773.6667) .. (148.0000,-779.0000) ..
    controls (202.0000,-828.3333) and (249.0000,-853.0000) .. (289.0000,-853.0000)
    .. controls (343.6667,-853.0000) and (404.3333,-778.6667) ..
    (471.0000,-630.0000) -- (507.0000,-550.0000) .. controls (755.6667,-996.0000)
    and (1074.6667,-1335.3333) .. (1464.0000,-1568.0000) -- (1494.0000,-1516.0000)
    -- cycle;
\end{scope}

\end{tikzpicture}} & \resizebox{!}{\fontcharht\font`M}{\begin{tikzpicture}[y=0.80pt, x=0.80pt, yscale=-1.000000, xscale=1.000000, inner sep=0pt, outer sep=0pt]
\begin{scope}[shift={(100.0,1731.0)},nonzero rule]
  \path[draw=.,fill=.,line width=1.600pt] (1271.0000,-1280.0000) ..
    controls (1271.0000,-1272.0000) and (1263.6667,-1261.0000) ..
    (1249.0000,-1247.0000) .. controls (1237.0000,-1235.6667) and
    (1213.6667,-1207.6667) .. (1179.0000,-1163.0000) -- (866.0000,-754.0000) --
    (929.0000,-627.0000) -- (1089.0000,-285.0000) .. controls
    (1083.0000,-271.0000) and (1073.6667,-261.3333) .. (1061.0000,-256.0000) --
    (1061.0000,-240.0000) .. controls (1060.3333,-239.3333) and
    (1052.3333,-234.3333) .. (1037.0000,-225.0000) .. controls
    (1025.6667,-217.6667) and (1020.0000,-211.6667) .. (1020.0000,-207.0000) ..
    controls (1020.0000,-201.0000) and (1023.3333,-191.3333) ..
    (1030.0000,-178.0000) .. controls (1028.0000,-166.0000) and
    (1019.3333,-154.3333) .. (1004.0000,-143.0000) .. controls
    (988.6667,-132.3333) and (974.6667,-127.0000) .. (962.0000,-127.0000) ..
    controls (948.6667,-127.0000) and (922.6667,-157.0000) .. (884.0000,-217.0000)
    -- (692.0000,-514.0000) -- (286.0000,115.0000) .. controls (264.6667,131.0000)
    and (229.3333,139.0000) .. (180.0000,139.0000) .. controls (168.6667,139.0000)
    and (154.3333,125.3333) .. (137.0000,98.0000) .. controls (116.3333,65.3333)
    and (106.0000,44.3333) .. (106.0000,35.0000) .. controls (106.0000,27.0000)
    and (108.0000,16.0000) .. (112.0000,2.0000) .. controls (118.0000,-18.6667)
    and (117.3333,-32.0000) .. (110.0000,-38.0000) .. controls (103.3333,-44.6667)
    and (100.0000,-49.6667) .. (100.0000,-53.0000) .. controls (100.0000,-64.3333)
    and (116.3333,-98.0000) .. (149.0000,-154.0000) -- (198.0000,-238.0000) ..
    controls (250.6667,-328.0000) and (367.6667,-482.6667) .. (549.0000,-702.0000)
    -- (573.0000,-731.0000) -- (422.0000,-1149.0000) .. controls
    (406.6667,-1191.6667) and (399.0000,-1215.0000) .. (399.0000,-1219.0000) ..
    controls (399.0000,-1225.0000) and (408.0000,-1239.3333) ..
    (426.0000,-1262.0000) -- (440.0000,-1274.0000) .. controls
    (444.6667,-1278.0000) and (449.3333,-1280.0000) .. (454.0000,-1280.0000) ..
    controls (461.3333,-1280.0000) and (472.3333,-1274.0000) ..
    (487.0000,-1262.0000) .. controls (495.0000,-1268.6667) and
    (506.0000,-1272.0000) .. (520.0000,-1272.0000) .. controls
    (524.0000,-1274.0000) and (530.0000,-1278.6667) .. (538.0000,-1286.0000) ..
    controls (544.6667,-1297.3333) and (552.3333,-1303.0000) ..
    (561.0000,-1303.0000) .. controls (573.6667,-1303.0000) and
    (586.0000,-1290.6667) .. (598.0000,-1266.0000) -- (753.0000,-958.0000) --
    (805.0000,-1022.0000) -- (905.0000,-1137.0000) .. controls
    (1029.6667,-1280.3333) and (1095.0000,-1352.0000) .. (1101.0000,-1352.0000) ..
    controls (1117.6667,-1352.0000) and (1141.6667,-1339.0000) ..
    (1173.0000,-1313.0000) -- (1196.0000,-1333.0000) .. controls
    (1205.3333,-1347.0000) and (1212.6667,-1354.0000) .. (1218.0000,-1354.0000) ..
    controls (1221.3333,-1354.0000) and (1225.3333,-1348.3333) ..
    (1230.0000,-1337.0000) .. controls (1232.6667,-1331.0000) and
    (1240.6667,-1320.6667) .. (1254.0000,-1306.0000) .. controls
    (1265.3333,-1294.0000) and (1271.0000,-1285.3333) .. (1271.0000,-1280.0000) --
    cycle;
\end{scope}

\end{tikzpicture}} & 3 & 4000 & 250×250  & 31.42M\\
				\hline
				SinDiffusion & \resizebox{!}{\fontcharht\font`M}{\begin{tikzpicture}[y=0.80pt, x=0.80pt, yscale=-1.000000, xscale=1.000000, inner sep=0pt, outer sep=0pt]
\begin{scope}[shift={(100.0,2042.0)},nonzero rule]
  \path[draw=.,fill=.,line width=1.600pt] (1494.0000,-1516.0000) ..
    controls (1303.3333,-1381.3333) and (1125.6667,-1211.3333) ..
    (961.0000,-1006.0000) .. controls (790.3333,-792.6667) and
    (666.6667,-581.0000) .. (590.0000,-371.0000) -- (546.0000,-342.0000) ..
    controls (506.6667,-316.0000) and (474.0000,-291.3333) .. (448.0000,-268.0000)
    .. controls (443.3333,-292.6667) and (430.3333,-331.3333) ..
    (409.0000,-384.0000) -- (386.0000,-441.0000) .. controls (347.3333,-537.0000)
    and (318.0000,-603.0000) .. (298.0000,-639.0000) .. controls
    (248.0000,-727.0000) and (198.0000,-773.6667) .. (148.0000,-779.0000) ..
    controls (202.0000,-828.3333) and (249.0000,-853.0000) .. (289.0000,-853.0000)
    .. controls (343.6667,-853.0000) and (404.3333,-778.6667) ..
    (471.0000,-630.0000) -- (507.0000,-550.0000) .. controls (755.6667,-996.0000)
    and (1074.6667,-1335.3333) .. (1464.0000,-1568.0000) -- (1494.0000,-1516.0000)
    -- cycle;
\end{scope}

\end{tikzpicture}} & \resizebox{!}{\fontcharht\font`M}{\begin{tikzpicture}[y=0.80pt, x=0.80pt, yscale=-1.000000, xscale=1.000000, inner sep=0pt, outer sep=0pt]
\begin{scope}[shift={(100.0,2042.0)},nonzero rule]
  \path[draw=.,fill=.,line width=1.600pt] (1494.0000,-1516.0000) ..
    controls (1303.3333,-1381.3333) and (1125.6667,-1211.3333) ..
    (961.0000,-1006.0000) .. controls (790.3333,-792.6667) and
    (666.6667,-581.0000) .. (590.0000,-371.0000) -- (546.0000,-342.0000) ..
    controls (506.6667,-316.0000) and (474.0000,-291.3333) .. (448.0000,-268.0000)
    .. controls (443.3333,-292.6667) and (430.3333,-331.3333) ..
    (409.0000,-384.0000) -- (386.0000,-441.0000) .. controls (347.3333,-537.0000)
    and (318.0000,-603.0000) .. (298.0000,-639.0000) .. controls
    (248.0000,-727.0000) and (198.0000,-773.6667) .. (148.0000,-779.0000) ..
    controls (202.0000,-828.3333) and (249.0000,-853.0000) .. (289.0000,-853.0000)
    .. controls (343.6667,-853.0000) and (404.3333,-778.6667) ..
    (471.0000,-630.0000) -- (507.0000,-550.0000) .. controls (755.6667,-996.0000)
    and (1074.6667,-1335.3333) .. (1464.0000,-1568.0000) -- (1494.0000,-1516.0000)
    -- cycle;
\end{scope}

\end{tikzpicture}} & \resizebox{!}{\fontcharht\font`M}{\begin{tikzpicture}[y=0.80pt, x=0.80pt, yscale=-1.000000, xscale=1.000000, inner sep=0pt, outer sep=0pt]
\begin{scope}[shift={(100.0,1731.0)},nonzero rule]
  \path[draw=.,fill=.,line width=1.600pt] (1271.0000,-1280.0000) ..
    controls (1271.0000,-1272.0000) and (1263.6667,-1261.0000) ..
    (1249.0000,-1247.0000) .. controls (1237.0000,-1235.6667) and
    (1213.6667,-1207.6667) .. (1179.0000,-1163.0000) -- (866.0000,-754.0000) --
    (929.0000,-627.0000) -- (1089.0000,-285.0000) .. controls
    (1083.0000,-271.0000) and (1073.6667,-261.3333) .. (1061.0000,-256.0000) --
    (1061.0000,-240.0000) .. controls (1060.3333,-239.3333) and
    (1052.3333,-234.3333) .. (1037.0000,-225.0000) .. controls
    (1025.6667,-217.6667) and (1020.0000,-211.6667) .. (1020.0000,-207.0000) ..
    controls (1020.0000,-201.0000) and (1023.3333,-191.3333) ..
    (1030.0000,-178.0000) .. controls (1028.0000,-166.0000) and
    (1019.3333,-154.3333) .. (1004.0000,-143.0000) .. controls
    (988.6667,-132.3333) and (974.6667,-127.0000) .. (962.0000,-127.0000) ..
    controls (948.6667,-127.0000) and (922.6667,-157.0000) .. (884.0000,-217.0000)
    -- (692.0000,-514.0000) -- (286.0000,115.0000) .. controls (264.6667,131.0000)
    and (229.3333,139.0000) .. (180.0000,139.0000) .. controls (168.6667,139.0000)
    and (154.3333,125.3333) .. (137.0000,98.0000) .. controls (116.3333,65.3333)
    and (106.0000,44.3333) .. (106.0000,35.0000) .. controls (106.0000,27.0000)
    and (108.0000,16.0000) .. (112.0000,2.0000) .. controls (118.0000,-18.6667)
    and (117.3333,-32.0000) .. (110.0000,-38.0000) .. controls (103.3333,-44.6667)
    and (100.0000,-49.6667) .. (100.0000,-53.0000) .. controls (100.0000,-64.3333)
    and (116.3333,-98.0000) .. (149.0000,-154.0000) -- (198.0000,-238.0000) ..
    controls (250.6667,-328.0000) and (367.6667,-482.6667) .. (549.0000,-702.0000)
    -- (573.0000,-731.0000) -- (422.0000,-1149.0000) .. controls
    (406.6667,-1191.6667) and (399.0000,-1215.0000) .. (399.0000,-1219.0000) ..
    controls (399.0000,-1225.0000) and (408.0000,-1239.3333) ..
    (426.0000,-1262.0000) -- (440.0000,-1274.0000) .. controls
    (444.6667,-1278.0000) and (449.3333,-1280.0000) .. (454.0000,-1280.0000) ..
    controls (461.3333,-1280.0000) and (472.3333,-1274.0000) ..
    (487.0000,-1262.0000) .. controls (495.0000,-1268.6667) and
    (506.0000,-1272.0000) .. (520.0000,-1272.0000) .. controls
    (524.0000,-1274.0000) and (530.0000,-1278.6667) .. (538.0000,-1286.0000) ..
    controls (544.6667,-1297.3333) and (552.3333,-1303.0000) ..
    (561.0000,-1303.0000) .. controls (573.6667,-1303.0000) and
    (586.0000,-1290.6667) .. (598.0000,-1266.0000) -- (753.0000,-958.0000) --
    (805.0000,-1022.0000) -- (905.0000,-1137.0000) .. controls
    (1029.6667,-1280.3333) and (1095.0000,-1352.0000) .. (1101.0000,-1352.0000) ..
    controls (1117.6667,-1352.0000) and (1141.6667,-1339.0000) ..
    (1173.0000,-1313.0000) -- (1196.0000,-1333.0000) .. controls
    (1205.3333,-1347.0000) and (1212.6667,-1354.0000) .. (1218.0000,-1354.0000) ..
    controls (1221.3333,-1354.0000) and (1225.3333,-1348.3333) ..
    (1230.0000,-1337.0000) .. controls (1232.6667,-1331.0000) and
    (1240.6667,-1320.6667) .. (1254.0000,-1306.0000) .. controls
    (1265.3333,-1294.0000) and (1271.0000,-1285.3333) .. (1271.0000,-1280.0000) --
    cycle;
\end{scope}

\end{tikzpicture}} & 1 & 1000 & 250×250 & 106.7M \\
				\hline
				TransGAN & \resizebox{!}{\fontcharht\font`M}{\begin{tikzpicture}[y=0.80pt, x=0.80pt, yscale=-1.000000, xscale=1.000000, inner sep=0pt, outer sep=0pt]
\begin{scope}[shift={(100.0,1731.0)},nonzero rule]
  \path[draw=.,fill=.,line width=1.600pt] (1271.0000,-1280.0000) ..
    controls (1271.0000,-1272.0000) and (1263.6667,-1261.0000) ..
    (1249.0000,-1247.0000) .. controls (1237.0000,-1235.6667) and
    (1213.6667,-1207.6667) .. (1179.0000,-1163.0000) -- (866.0000,-754.0000) --
    (929.0000,-627.0000) -- (1089.0000,-285.0000) .. controls
    (1083.0000,-271.0000) and (1073.6667,-261.3333) .. (1061.0000,-256.0000) --
    (1061.0000,-240.0000) .. controls (1060.3333,-239.3333) and
    (1052.3333,-234.3333) .. (1037.0000,-225.0000) .. controls
    (1025.6667,-217.6667) and (1020.0000,-211.6667) .. (1020.0000,-207.0000) ..
    controls (1020.0000,-201.0000) and (1023.3333,-191.3333) ..
    (1030.0000,-178.0000) .. controls (1028.0000,-166.0000) and
    (1019.3333,-154.3333) .. (1004.0000,-143.0000) .. controls
    (988.6667,-132.3333) and (974.6667,-127.0000) .. (962.0000,-127.0000) ..
    controls (948.6667,-127.0000) and (922.6667,-157.0000) .. (884.0000,-217.0000)
    -- (692.0000,-514.0000) -- (286.0000,115.0000) .. controls (264.6667,131.0000)
    and (229.3333,139.0000) .. (180.0000,139.0000) .. controls (168.6667,139.0000)
    and (154.3333,125.3333) .. (137.0000,98.0000) .. controls (116.3333,65.3333)
    and (106.0000,44.3333) .. (106.0000,35.0000) .. controls (106.0000,27.0000)
    and (108.0000,16.0000) .. (112.0000,2.0000) .. controls (118.0000,-18.6667)
    and (117.3333,-32.0000) .. (110.0000,-38.0000) .. controls (103.3333,-44.6667)
    and (100.0000,-49.6667) .. (100.0000,-53.0000) .. controls (100.0000,-64.3333)
    and (116.3333,-98.0000) .. (149.0000,-154.0000) -- (198.0000,-238.0000) ..
    controls (250.6667,-328.0000) and (367.6667,-482.6667) .. (549.0000,-702.0000)
    -- (573.0000,-731.0000) -- (422.0000,-1149.0000) .. controls
    (406.6667,-1191.6667) and (399.0000,-1215.0000) .. (399.0000,-1219.0000) ..
    controls (399.0000,-1225.0000) and (408.0000,-1239.3333) ..
    (426.0000,-1262.0000) -- (440.0000,-1274.0000) .. controls
    (444.6667,-1278.0000) and (449.3333,-1280.0000) .. (454.0000,-1280.0000) ..
    controls (461.3333,-1280.0000) and (472.3333,-1274.0000) ..
    (487.0000,-1262.0000) .. controls (495.0000,-1268.6667) and
    (506.0000,-1272.0000) .. (520.0000,-1272.0000) .. controls
    (524.0000,-1274.0000) and (530.0000,-1278.6667) .. (538.0000,-1286.0000) ..
    controls (544.6667,-1297.3333) and (552.3333,-1303.0000) ..
    (561.0000,-1303.0000) .. controls (573.6667,-1303.0000) and
    (586.0000,-1290.6667) .. (598.0000,-1266.0000) -- (753.0000,-958.0000) --
    (805.0000,-1022.0000) -- (905.0000,-1137.0000) .. controls
    (1029.6667,-1280.3333) and (1095.0000,-1352.0000) .. (1101.0000,-1352.0000) ..
    controls (1117.6667,-1352.0000) and (1141.6667,-1339.0000) ..
    (1173.0000,-1313.0000) -- (1196.0000,-1333.0000) .. controls
    (1205.3333,-1347.0000) and (1212.6667,-1354.0000) .. (1218.0000,-1354.0000) ..
    controls (1221.3333,-1354.0000) and (1225.3333,-1348.3333) ..
    (1230.0000,-1337.0000) .. controls (1232.6667,-1331.0000) and
    (1240.6667,-1320.6667) .. (1254.0000,-1306.0000) .. controls
    (1265.3333,-1294.0000) and (1271.0000,-1285.3333) .. (1271.0000,-1280.0000) --
    cycle;
\end{scope}

\end{tikzpicture}} & \resizebox{!}{\fontcharht\font`M}{\begin{tikzpicture}[y=0.80pt, x=0.80pt, yscale=-1.000000, xscale=1.000000, inner sep=0pt, outer sep=0pt]
\begin{scope}[shift={(100.0,1731.0)},nonzero rule]
  \path[draw=.,fill=.,line width=1.600pt] (1271.0000,-1280.0000) ..
    controls (1271.0000,-1272.0000) and (1263.6667,-1261.0000) ..
    (1249.0000,-1247.0000) .. controls (1237.0000,-1235.6667) and
    (1213.6667,-1207.6667) .. (1179.0000,-1163.0000) -- (866.0000,-754.0000) --
    (929.0000,-627.0000) -- (1089.0000,-285.0000) .. controls
    (1083.0000,-271.0000) and (1073.6667,-261.3333) .. (1061.0000,-256.0000) --
    (1061.0000,-240.0000) .. controls (1060.3333,-239.3333) and
    (1052.3333,-234.3333) .. (1037.0000,-225.0000) .. controls
    (1025.6667,-217.6667) and (1020.0000,-211.6667) .. (1020.0000,-207.0000) ..
    controls (1020.0000,-201.0000) and (1023.3333,-191.3333) ..
    (1030.0000,-178.0000) .. controls (1028.0000,-166.0000) and
    (1019.3333,-154.3333) .. (1004.0000,-143.0000) .. controls
    (988.6667,-132.3333) and (974.6667,-127.0000) .. (962.0000,-127.0000) ..
    controls (948.6667,-127.0000) and (922.6667,-157.0000) .. (884.0000,-217.0000)
    -- (692.0000,-514.0000) -- (286.0000,115.0000) .. controls (264.6667,131.0000)
    and (229.3333,139.0000) .. (180.0000,139.0000) .. controls (168.6667,139.0000)
    and (154.3333,125.3333) .. (137.0000,98.0000) .. controls (116.3333,65.3333)
    and (106.0000,44.3333) .. (106.0000,35.0000) .. controls (106.0000,27.0000)
    and (108.0000,16.0000) .. (112.0000,2.0000) .. controls (118.0000,-18.6667)
    and (117.3333,-32.0000) .. (110.0000,-38.0000) .. controls (103.3333,-44.6667)
    and (100.0000,-49.6667) .. (100.0000,-53.0000) .. controls (100.0000,-64.3333)
    and (116.3333,-98.0000) .. (149.0000,-154.0000) -- (198.0000,-238.0000) ..
    controls (250.6667,-328.0000) and (367.6667,-482.6667) .. (549.0000,-702.0000)
    -- (573.0000,-731.0000) -- (422.0000,-1149.0000) .. controls
    (406.6667,-1191.6667) and (399.0000,-1215.0000) .. (399.0000,-1219.0000) ..
    controls (399.0000,-1225.0000) and (408.0000,-1239.3333) ..
    (426.0000,-1262.0000) -- (440.0000,-1274.0000) .. controls
    (444.6667,-1278.0000) and (449.3333,-1280.0000) .. (454.0000,-1280.0000) ..
    controls (461.3333,-1280.0000) and (472.3333,-1274.0000) ..
    (487.0000,-1262.0000) .. controls (495.0000,-1268.6667) and
    (506.0000,-1272.0000) .. (520.0000,-1272.0000) .. controls
    (524.0000,-1274.0000) and (530.0000,-1278.6667) .. (538.0000,-1286.0000) ..
    controls (544.6667,-1297.3333) and (552.3333,-1303.0000) ..
    (561.0000,-1303.0000) .. controls (573.6667,-1303.0000) and
    (586.0000,-1290.6667) .. (598.0000,-1266.0000) -- (753.0000,-958.0000) --
    (805.0000,-1022.0000) -- (905.0000,-1137.0000) .. controls
    (1029.6667,-1280.3333) and (1095.0000,-1352.0000) .. (1101.0000,-1352.0000) ..
    controls (1117.6667,-1352.0000) and (1141.6667,-1339.0000) ..
    (1173.0000,-1313.0000) -- (1196.0000,-1333.0000) .. controls
    (1205.3333,-1347.0000) and (1212.6667,-1354.0000) .. (1218.0000,-1354.0000) ..
    controls (1221.3333,-1354.0000) and (1225.3333,-1348.3333) ..
    (1230.0000,-1337.0000) .. controls (1232.6667,-1331.0000) and
    (1240.6667,-1320.6667) .. (1254.0000,-1306.0000) .. controls
    (1265.3333,-1294.0000) and (1271.0000,-1285.3333) .. (1271.0000,-1280.0000) --
    cycle;
\end{scope}

\end{tikzpicture}} & \resizebox{!}{\fontcharht\font`M}{\begin{tikzpicture}[y=0.80pt, x=0.80pt, yscale=-1.000000, xscale=1.000000, inner sep=0pt, outer sep=0pt]
\begin{scope}[shift={(100.0,2042.0)},nonzero rule]
  \path[draw=.,fill=.,line width=1.600pt] (1494.0000,-1516.0000) ..
    controls (1303.3333,-1381.3333) and (1125.6667,-1211.3333) ..
    (961.0000,-1006.0000) .. controls (790.3333,-792.6667) and
    (666.6667,-581.0000) .. (590.0000,-371.0000) -- (546.0000,-342.0000) ..
    controls (506.6667,-316.0000) and (474.0000,-291.3333) .. (448.0000,-268.0000)
    .. controls (443.3333,-292.6667) and (430.3333,-331.3333) ..
    (409.0000,-384.0000) -- (386.0000,-441.0000) .. controls (347.3333,-537.0000)
    and (318.0000,-603.0000) .. (298.0000,-639.0000) .. controls
    (248.0000,-727.0000) and (198.0000,-773.6667) .. (148.0000,-779.0000) ..
    controls (202.0000,-828.3333) and (249.0000,-853.0000) .. (289.0000,-853.0000)
    .. controls (343.6667,-853.0000) and (404.3333,-778.6667) ..
    (471.0000,-630.0000) -- (507.0000,-550.0000) .. controls (755.6667,-996.0000)
    and (1074.6667,-1335.3333) .. (1464.0000,-1568.0000) -- (1494.0000,-1516.0000)
    -- cycle;
\end{scope}

\end{tikzpicture}} & 1 & 500000 & 64×64 & \resizebox{!}{\fontcharht\font`M}{\begin{tikzpicture}[y=0.80pt, x=0.80pt, yscale=-1.000000, xscale=1.000000, inner sep=0pt, outer sep=0pt]
\begin{scope}[shift={(100.0,1731.0)},nonzero rule]
  \path[draw=.,fill=.,line width=1.600pt] (1271.0000,-1280.0000) ..
    controls (1271.0000,-1272.0000) and (1263.6667,-1261.0000) ..
    (1249.0000,-1247.0000) .. controls (1237.0000,-1235.6667) and
    (1213.6667,-1207.6667) .. (1179.0000,-1163.0000) -- (866.0000,-754.0000) --
    (929.0000,-627.0000) -- (1089.0000,-285.0000) .. controls
    (1083.0000,-271.0000) and (1073.6667,-261.3333) .. (1061.0000,-256.0000) --
    (1061.0000,-240.0000) .. controls (1060.3333,-239.3333) and
    (1052.3333,-234.3333) .. (1037.0000,-225.0000) .. controls
    (1025.6667,-217.6667) and (1020.0000,-211.6667) .. (1020.0000,-207.0000) ..
    controls (1020.0000,-201.0000) and (1023.3333,-191.3333) ..
    (1030.0000,-178.0000) .. controls (1028.0000,-166.0000) and
    (1019.3333,-154.3333) .. (1004.0000,-143.0000) .. controls
    (988.6667,-132.3333) and (974.6667,-127.0000) .. (962.0000,-127.0000) ..
    controls (948.6667,-127.0000) and (922.6667,-157.0000) .. (884.0000,-217.0000)
    -- (692.0000,-514.0000) -- (286.0000,115.0000) .. controls (264.6667,131.0000)
    and (229.3333,139.0000) .. (180.0000,139.0000) .. controls (168.6667,139.0000)
    and (154.3333,125.3333) .. (137.0000,98.0000) .. controls (116.3333,65.3333)
    and (106.0000,44.3333) .. (106.0000,35.0000) .. controls (106.0000,27.0000)
    and (108.0000,16.0000) .. (112.0000,2.0000) .. controls (118.0000,-18.6667)
    and (117.3333,-32.0000) .. (110.0000,-38.0000) .. controls (103.3333,-44.6667)
    and (100.0000,-49.6667) .. (100.0000,-53.0000) .. controls (100.0000,-64.3333)
    and (116.3333,-98.0000) .. (149.0000,-154.0000) -- (198.0000,-238.0000) ..
    controls (250.6667,-328.0000) and (367.6667,-482.6667) .. (549.0000,-702.0000)
    -- (573.0000,-731.0000) -- (422.0000,-1149.0000) .. controls
    (406.6667,-1191.6667) and (399.0000,-1215.0000) .. (399.0000,-1219.0000) ..
    controls (399.0000,-1225.0000) and (408.0000,-1239.3333) ..
    (426.0000,-1262.0000) -- (440.0000,-1274.0000) .. controls
    (444.6667,-1278.0000) and (449.3333,-1280.0000) .. (454.0000,-1280.0000) ..
    controls (461.3333,-1280.0000) and (472.3333,-1274.0000) ..
    (487.0000,-1262.0000) .. controls (495.0000,-1268.6667) and
    (506.0000,-1272.0000) .. (520.0000,-1272.0000) .. controls
    (524.0000,-1274.0000) and (530.0000,-1278.6667) .. (538.0000,-1286.0000) ..
    controls (544.6667,-1297.3333) and (552.3333,-1303.0000) ..
    (561.0000,-1303.0000) .. controls (573.6667,-1303.0000) and
    (586.0000,-1290.6667) .. (598.0000,-1266.0000) -- (753.0000,-958.0000) --
    (805.0000,-1022.0000) -- (905.0000,-1137.0000) .. controls
    (1029.6667,-1280.3333) and (1095.0000,-1352.0000) .. (1101.0000,-1352.0000) ..
    controls (1117.6667,-1352.0000) and (1141.6667,-1339.0000) ..
    (1173.0000,-1313.0000) -- (1196.0000,-1333.0000) .. controls
    (1205.3333,-1347.0000) and (1212.6667,-1354.0000) .. (1218.0000,-1354.0000) ..
    controls (1221.3333,-1354.0000) and (1225.3333,-1348.3333) ..
    (1230.0000,-1337.0000) .. controls (1232.6667,-1331.0000) and
    (1240.6667,-1320.6667) .. (1254.0000,-1306.0000) .. controls
    (1265.3333,-1294.0000) and (1271.0000,-1285.3333) .. (1271.0000,-1280.0000) --
    cycle;
\end{scope}

\end{tikzpicture}} \\
				\hline
				ViTGAN & \resizebox{!}{\fontcharht\font`M}{\begin{tikzpicture}[y=0.80pt, x=0.80pt, yscale=-1.000000, xscale=1.000000, inner sep=0pt, outer sep=0pt]
\begin{scope}[shift={(100.0,1731.0)},nonzero rule]
  \path[draw=.,fill=.,line width=1.600pt] (1271.0000,-1280.0000) ..
    controls (1271.0000,-1272.0000) and (1263.6667,-1261.0000) ..
    (1249.0000,-1247.0000) .. controls (1237.0000,-1235.6667) and
    (1213.6667,-1207.6667) .. (1179.0000,-1163.0000) -- (866.0000,-754.0000) --
    (929.0000,-627.0000) -- (1089.0000,-285.0000) .. controls
    (1083.0000,-271.0000) and (1073.6667,-261.3333) .. (1061.0000,-256.0000) --
    (1061.0000,-240.0000) .. controls (1060.3333,-239.3333) and
    (1052.3333,-234.3333) .. (1037.0000,-225.0000) .. controls
    (1025.6667,-217.6667) and (1020.0000,-211.6667) .. (1020.0000,-207.0000) ..
    controls (1020.0000,-201.0000) and (1023.3333,-191.3333) ..
    (1030.0000,-178.0000) .. controls (1028.0000,-166.0000) and
    (1019.3333,-154.3333) .. (1004.0000,-143.0000) .. controls
    (988.6667,-132.3333) and (974.6667,-127.0000) .. (962.0000,-127.0000) ..
    controls (948.6667,-127.0000) and (922.6667,-157.0000) .. (884.0000,-217.0000)
    -- (692.0000,-514.0000) -- (286.0000,115.0000) .. controls (264.6667,131.0000)
    and (229.3333,139.0000) .. (180.0000,139.0000) .. controls (168.6667,139.0000)
    and (154.3333,125.3333) .. (137.0000,98.0000) .. controls (116.3333,65.3333)
    and (106.0000,44.3333) .. (106.0000,35.0000) .. controls (106.0000,27.0000)
    and (108.0000,16.0000) .. (112.0000,2.0000) .. controls (118.0000,-18.6667)
    and (117.3333,-32.0000) .. (110.0000,-38.0000) .. controls (103.3333,-44.6667)
    and (100.0000,-49.6667) .. (100.0000,-53.0000) .. controls (100.0000,-64.3333)
    and (116.3333,-98.0000) .. (149.0000,-154.0000) -- (198.0000,-238.0000) ..
    controls (250.6667,-328.0000) and (367.6667,-482.6667) .. (549.0000,-702.0000)
    -- (573.0000,-731.0000) -- (422.0000,-1149.0000) .. controls
    (406.6667,-1191.6667) and (399.0000,-1215.0000) .. (399.0000,-1219.0000) ..
    controls (399.0000,-1225.0000) and (408.0000,-1239.3333) ..
    (426.0000,-1262.0000) -- (440.0000,-1274.0000) .. controls
    (444.6667,-1278.0000) and (449.3333,-1280.0000) .. (454.0000,-1280.0000) ..
    controls (461.3333,-1280.0000) and (472.3333,-1274.0000) ..
    (487.0000,-1262.0000) .. controls (495.0000,-1268.6667) and
    (506.0000,-1272.0000) .. (520.0000,-1272.0000) .. controls
    (524.0000,-1274.0000) and (530.0000,-1278.6667) .. (538.0000,-1286.0000) ..
    controls (544.6667,-1297.3333) and (552.3333,-1303.0000) ..
    (561.0000,-1303.0000) .. controls (573.6667,-1303.0000) and
    (586.0000,-1290.6667) .. (598.0000,-1266.0000) -- (753.0000,-958.0000) --
    (805.0000,-1022.0000) -- (905.0000,-1137.0000) .. controls
    (1029.6667,-1280.3333) and (1095.0000,-1352.0000) .. (1101.0000,-1352.0000) ..
    controls (1117.6667,-1352.0000) and (1141.6667,-1339.0000) ..
    (1173.0000,-1313.0000) -- (1196.0000,-1333.0000) .. controls
    (1205.3333,-1347.0000) and (1212.6667,-1354.0000) .. (1218.0000,-1354.0000) ..
    controls (1221.3333,-1354.0000) and (1225.3333,-1348.3333) ..
    (1230.0000,-1337.0000) .. controls (1232.6667,-1331.0000) and
    (1240.6667,-1320.6667) .. (1254.0000,-1306.0000) .. controls
    (1265.3333,-1294.0000) and (1271.0000,-1285.3333) .. (1271.0000,-1280.0000) --
    cycle;
\end{scope}

\end{tikzpicture}} & \resizebox{!}{\fontcharht\font`M}{\begin{tikzpicture}[y=0.80pt, x=0.80pt, yscale=-1.000000, xscale=1.000000, inner sep=0pt, outer sep=0pt]
\begin{scope}[shift={(100.0,1731.0)},nonzero rule]
  \path[draw=.,fill=.,line width=1.600pt] (1271.0000,-1280.0000) ..
    controls (1271.0000,-1272.0000) and (1263.6667,-1261.0000) ..
    (1249.0000,-1247.0000) .. controls (1237.0000,-1235.6667) and
    (1213.6667,-1207.6667) .. (1179.0000,-1163.0000) -- (866.0000,-754.0000) --
    (929.0000,-627.0000) -- (1089.0000,-285.0000) .. controls
    (1083.0000,-271.0000) and (1073.6667,-261.3333) .. (1061.0000,-256.0000) --
    (1061.0000,-240.0000) .. controls (1060.3333,-239.3333) and
    (1052.3333,-234.3333) .. (1037.0000,-225.0000) .. controls
    (1025.6667,-217.6667) and (1020.0000,-211.6667) .. (1020.0000,-207.0000) ..
    controls (1020.0000,-201.0000) and (1023.3333,-191.3333) ..
    (1030.0000,-178.0000) .. controls (1028.0000,-166.0000) and
    (1019.3333,-154.3333) .. (1004.0000,-143.0000) .. controls
    (988.6667,-132.3333) and (974.6667,-127.0000) .. (962.0000,-127.0000) ..
    controls (948.6667,-127.0000) and (922.6667,-157.0000) .. (884.0000,-217.0000)
    -- (692.0000,-514.0000) -- (286.0000,115.0000) .. controls (264.6667,131.0000)
    and (229.3333,139.0000) .. (180.0000,139.0000) .. controls (168.6667,139.0000)
    and (154.3333,125.3333) .. (137.0000,98.0000) .. controls (116.3333,65.3333)
    and (106.0000,44.3333) .. (106.0000,35.0000) .. controls (106.0000,27.0000)
    and (108.0000,16.0000) .. (112.0000,2.0000) .. controls (118.0000,-18.6667)
    and (117.3333,-32.0000) .. (110.0000,-38.0000) .. controls (103.3333,-44.6667)
    and (100.0000,-49.6667) .. (100.0000,-53.0000) .. controls (100.0000,-64.3333)
    and (116.3333,-98.0000) .. (149.0000,-154.0000) -- (198.0000,-238.0000) ..
    controls (250.6667,-328.0000) and (367.6667,-482.6667) .. (549.0000,-702.0000)
    -- (573.0000,-731.0000) -- (422.0000,-1149.0000) .. controls
    (406.6667,-1191.6667) and (399.0000,-1215.0000) .. (399.0000,-1219.0000) ..
    controls (399.0000,-1225.0000) and (408.0000,-1239.3333) ..
    (426.0000,-1262.0000) -- (440.0000,-1274.0000) .. controls
    (444.6667,-1278.0000) and (449.3333,-1280.0000) .. (454.0000,-1280.0000) ..
    controls (461.3333,-1280.0000) and (472.3333,-1274.0000) ..
    (487.0000,-1262.0000) .. controls (495.0000,-1268.6667) and
    (506.0000,-1272.0000) .. (520.0000,-1272.0000) .. controls
    (524.0000,-1274.0000) and (530.0000,-1278.6667) .. (538.0000,-1286.0000) ..
    controls (544.6667,-1297.3333) and (552.3333,-1303.0000) ..
    (561.0000,-1303.0000) .. controls (573.6667,-1303.0000) and
    (586.0000,-1290.6667) .. (598.0000,-1266.0000) -- (753.0000,-958.0000) --
    (805.0000,-1022.0000) -- (905.0000,-1137.0000) .. controls
    (1029.6667,-1280.3333) and (1095.0000,-1352.0000) .. (1101.0000,-1352.0000) ..
    controls (1117.6667,-1352.0000) and (1141.6667,-1339.0000) ..
    (1173.0000,-1313.0000) -- (1196.0000,-1333.0000) .. controls
    (1205.3333,-1347.0000) and (1212.6667,-1354.0000) .. (1218.0000,-1354.0000) ..
    controls (1221.3333,-1354.0000) and (1225.3333,-1348.3333) ..
    (1230.0000,-1337.0000) .. controls (1232.6667,-1331.0000) and
    (1240.6667,-1320.6667) .. (1254.0000,-1306.0000) .. controls
    (1265.3333,-1294.0000) and (1271.0000,-1285.3333) .. (1271.0000,-1280.0000) --
    cycle;
\end{scope}

\end{tikzpicture}} & \resizebox{!}{\fontcharht\font`M}{\begin{tikzpicture}[y=0.80pt, x=0.80pt, yscale=-1.000000, xscale=1.000000, inner sep=0pt, outer sep=0pt]
\begin{scope}[shift={(100.0,2042.0)},nonzero rule]
  \path[draw=.,fill=.,line width=1.600pt] (1494.0000,-1516.0000) ..
    controls (1303.3333,-1381.3333) and (1125.6667,-1211.3333) ..
    (961.0000,-1006.0000) .. controls (790.3333,-792.6667) and
    (666.6667,-581.0000) .. (590.0000,-371.0000) -- (546.0000,-342.0000) ..
    controls (506.6667,-316.0000) and (474.0000,-291.3333) .. (448.0000,-268.0000)
    .. controls (443.3333,-292.6667) and (430.3333,-331.3333) ..
    (409.0000,-384.0000) -- (386.0000,-441.0000) .. controls (347.3333,-537.0000)
    and (318.0000,-603.0000) .. (298.0000,-639.0000) .. controls
    (248.0000,-727.0000) and (198.0000,-773.6667) .. (148.0000,-779.0000) ..
    controls (202.0000,-828.3333) and (249.0000,-853.0000) .. (289.0000,-853.0000)
    .. controls (343.6667,-853.0000) and (404.3333,-778.6667) ..
    (471.0000,-630.0000) -- (507.0000,-550.0000) .. controls (755.6667,-996.0000)
    and (1074.6667,-1335.3333) .. (1464.0000,-1568.0000) -- (1494.0000,-1516.0000)
    -- cycle;
\end{scope}

\end{tikzpicture}} & 1 & 100000  & 64×64 & \resizebox{!}{\fontcharht\font`M}{\begin{tikzpicture}[y=0.80pt, x=0.80pt, yscale=-1.000000, xscale=1.000000, inner sep=0pt, outer sep=0pt]
\begin{scope}[shift={(100.0,1731.0)},nonzero rule]
  \path[draw=.,fill=.,line width=1.600pt] (1271.0000,-1280.0000) ..
    controls (1271.0000,-1272.0000) and (1263.6667,-1261.0000) ..
    (1249.0000,-1247.0000) .. controls (1237.0000,-1235.6667) and
    (1213.6667,-1207.6667) .. (1179.0000,-1163.0000) -- (866.0000,-754.0000) --
    (929.0000,-627.0000) -- (1089.0000,-285.0000) .. controls
    (1083.0000,-271.0000) and (1073.6667,-261.3333) .. (1061.0000,-256.0000) --
    (1061.0000,-240.0000) .. controls (1060.3333,-239.3333) and
    (1052.3333,-234.3333) .. (1037.0000,-225.0000) .. controls
    (1025.6667,-217.6667) and (1020.0000,-211.6667) .. (1020.0000,-207.0000) ..
    controls (1020.0000,-201.0000) and (1023.3333,-191.3333) ..
    (1030.0000,-178.0000) .. controls (1028.0000,-166.0000) and
    (1019.3333,-154.3333) .. (1004.0000,-143.0000) .. controls
    (988.6667,-132.3333) and (974.6667,-127.0000) .. (962.0000,-127.0000) ..
    controls (948.6667,-127.0000) and (922.6667,-157.0000) .. (884.0000,-217.0000)
    -- (692.0000,-514.0000) -- (286.0000,115.0000) .. controls (264.6667,131.0000)
    and (229.3333,139.0000) .. (180.0000,139.0000) .. controls (168.6667,139.0000)
    and (154.3333,125.3333) .. (137.0000,98.0000) .. controls (116.3333,65.3333)
    and (106.0000,44.3333) .. (106.0000,35.0000) .. controls (106.0000,27.0000)
    and (108.0000,16.0000) .. (112.0000,2.0000) .. controls (118.0000,-18.6667)
    and (117.3333,-32.0000) .. (110.0000,-38.0000) .. controls (103.3333,-44.6667)
    and (100.0000,-49.6667) .. (100.0000,-53.0000) .. controls (100.0000,-64.3333)
    and (116.3333,-98.0000) .. (149.0000,-154.0000) -- (198.0000,-238.0000) ..
    controls (250.6667,-328.0000) and (367.6667,-482.6667) .. (549.0000,-702.0000)
    -- (573.0000,-731.0000) -- (422.0000,-1149.0000) .. controls
    (406.6667,-1191.6667) and (399.0000,-1215.0000) .. (399.0000,-1219.0000) ..
    controls (399.0000,-1225.0000) and (408.0000,-1239.3333) ..
    (426.0000,-1262.0000) -- (440.0000,-1274.0000) .. controls
    (444.6667,-1278.0000) and (449.3333,-1280.0000) .. (454.0000,-1280.0000) ..
    controls (461.3333,-1280.0000) and (472.3333,-1274.0000) ..
    (487.0000,-1262.0000) .. controls (495.0000,-1268.6667) and
    (506.0000,-1272.0000) .. (520.0000,-1272.0000) .. controls
    (524.0000,-1274.0000) and (530.0000,-1278.6667) .. (538.0000,-1286.0000) ..
    controls (544.6667,-1297.3333) and (552.3333,-1303.0000) ..
    (561.0000,-1303.0000) .. controls (573.6667,-1303.0000) and
    (586.0000,-1290.6667) .. (598.0000,-1266.0000) -- (753.0000,-958.0000) --
    (805.0000,-1022.0000) -- (905.0000,-1137.0000) .. controls
    (1029.6667,-1280.3333) and (1095.0000,-1352.0000) .. (1101.0000,-1352.0000) ..
    controls (1117.6667,-1352.0000) and (1141.6667,-1339.0000) ..
    (1173.0000,-1313.0000) -- (1196.0000,-1333.0000) .. controls
    (1205.3333,-1347.0000) and (1212.6667,-1354.0000) .. (1218.0000,-1354.0000) ..
    controls (1221.3333,-1354.0000) and (1225.3333,-1348.3333) ..
    (1230.0000,-1337.0000) .. controls (1232.6667,-1331.0000) and
    (1240.6667,-1320.6667) .. (1254.0000,-1306.0000) .. controls
    (1265.3333,-1294.0000) and (1271.0000,-1285.3333) .. (1271.0000,-1280.0000) --
    cycle;
\end{scope}

\end{tikzpicture}} \\
				\hline
				\hline
				TcGAN & \resizebox{!}{\fontcharht\font`M}{\begin{tikzpicture}[y=0.80pt, x=0.80pt, yscale=-1.000000, xscale=1.000000, inner sep=0pt, outer sep=0pt]
\begin{scope}[shift={(100.0,2042.0)},nonzero rule]
  \path[draw=.,fill=.,line width=1.600pt] (1494.0000,-1516.0000) ..
    controls (1303.3333,-1381.3333) and (1125.6667,-1211.3333) ..
    (961.0000,-1006.0000) .. controls (790.3333,-792.6667) and
    (666.6667,-581.0000) .. (590.0000,-371.0000) -- (546.0000,-342.0000) ..
    controls (506.6667,-316.0000) and (474.0000,-291.3333) .. (448.0000,-268.0000)
    .. controls (443.3333,-292.6667) and (430.3333,-331.3333) ..
    (409.0000,-384.0000) -- (386.0000,-441.0000) .. controls (347.3333,-537.0000)
    and (318.0000,-603.0000) .. (298.0000,-639.0000) .. controls
    (248.0000,-727.0000) and (198.0000,-773.6667) .. (148.0000,-779.0000) ..
    controls (202.0000,-828.3333) and (249.0000,-853.0000) .. (289.0000,-853.0000)
    .. controls (343.6667,-853.0000) and (404.3333,-778.6667) ..
    (471.0000,-630.0000) -- (507.0000,-550.0000) .. controls (755.6667,-996.0000)
    and (1074.6667,-1335.3333) .. (1464.0000,-1568.0000) -- (1494.0000,-1516.0000)
    -- cycle;
\end{scope}

\end{tikzpicture}} & \resizebox{!}{\fontcharht\font`M}{\begin{tikzpicture}[y=0.80pt, x=0.80pt, yscale=-1.000000, xscale=1.000000, inner sep=0pt, outer sep=0pt]
\begin{scope}[shift={(100.0,2042.0)},nonzero rule]
  \path[draw=.,fill=.,line width=1.600pt] (1494.0000,-1516.0000) ..
    controls (1303.3333,-1381.3333) and (1125.6667,-1211.3333) ..
    (961.0000,-1006.0000) .. controls (790.3333,-792.6667) and
    (666.6667,-581.0000) .. (590.0000,-371.0000) -- (546.0000,-342.0000) ..
    controls (506.6667,-316.0000) and (474.0000,-291.3333) .. (448.0000,-268.0000)
    .. controls (443.3333,-292.6667) and (430.3333,-331.3333) ..
    (409.0000,-384.0000) -- (386.0000,-441.0000) .. controls (347.3333,-537.0000)
    and (318.0000,-603.0000) .. (298.0000,-639.0000) .. controls
    (248.0000,-727.0000) and (198.0000,-773.6667) .. (148.0000,-779.0000) ..
    controls (202.0000,-828.3333) and (249.0000,-853.0000) .. (289.0000,-853.0000)
    .. controls (343.6667,-853.0000) and (404.3333,-778.6667) ..
    (471.0000,-630.0000) -- (507.0000,-550.0000) .. controls (755.6667,-996.0000)
    and (1074.6667,-1335.3333) .. (1464.0000,-1568.0000) -- (1494.0000,-1516.0000)
    -- cycle;
\end{scope}

\end{tikzpicture}} & \resizebox{!}{\fontcharht\font`M}{\begin{tikzpicture}[y=0.80pt, x=0.80pt, yscale=-1.000000, xscale=1.000000, inner sep=0pt, outer sep=0pt]
\begin{scope}[shift={(100.0,2042.0)},nonzero rule]
  \path[draw=.,fill=.,line width=1.600pt] (1494.0000,-1516.0000) ..
    controls (1303.3333,-1381.3333) and (1125.6667,-1211.3333) ..
    (961.0000,-1006.0000) .. controls (790.3333,-792.6667) and
    (666.6667,-581.0000) .. (590.0000,-371.0000) -- (546.0000,-342.0000) ..
    controls (506.6667,-316.0000) and (474.0000,-291.3333) .. (448.0000,-268.0000)
    .. controls (443.3333,-292.6667) and (430.3333,-331.3333) ..
    (409.0000,-384.0000) -- (386.0000,-441.0000) .. controls (347.3333,-537.0000)
    and (318.0000,-603.0000) .. (298.0000,-639.0000) .. controls
    (248.0000,-727.0000) and (198.0000,-773.6667) .. (148.0000,-779.0000) ..
    controls (202.0000,-828.3333) and (249.0000,-853.0000) .. (289.0000,-853.0000)
    .. controls (343.6667,-853.0000) and (404.3333,-778.6667) ..
    (471.0000,-630.0000) -- (507.0000,-550.0000) .. controls (755.6667,-996.0000)
    and (1074.6667,-1335.3333) .. (1464.0000,-1568.0000) -- (1494.0000,-1516.0000)
    -- cycle;
\end{scope}

\end{tikzpicture}} & 6 & 1000  & 250×250 & 11.29M\\
				\hline 
			\end{tabular}
		}
	\end{center}
\end{table}

\noindent {\bf{Implementations Details.}} \quad Our method TcGAN is trained on 1 NVIDIA TU102 [TITAN RTX]. During training, all training images were randomly sampled and cropped to a fixed resolution of 256×256. The training images were scaled with a minimum resolution fixed at 25×25 (denoted as $ min $) and a maximum resolution of 250×250 (denoted as $ max $). Parameters represent the parameter scale of the model durinig training. For each training stage, we set the number of iterations (denoted as $ iter $) to 1000,  the number of $ G_T $ (denoted as $ n $) to 1, the number of stages (denoted as $ N $) to 6. In Table \ref{tab1}, we use \resizebox{!}{\fontcharht\font`M}{\begin{tikzpicture}[y=0.80pt, x=0.80pt, yscale=-1.000000, xscale=1.000000, inner sep=0pt, outer sep=0pt]
\begin{scope}[shift={(100.0,2042.0)},nonzero rule]
  \path[draw=.,fill=.,line width=1.600pt] (1494.0000,-1516.0000) ..
    controls (1303.3333,-1381.3333) and (1125.6667,-1211.3333) ..
    (961.0000,-1006.0000) .. controls (790.3333,-792.6667) and
    (666.6667,-581.0000) .. (590.0000,-371.0000) -- (546.0000,-342.0000) ..
    controls (506.6667,-316.0000) and (474.0000,-291.3333) .. (448.0000,-268.0000)
    .. controls (443.3333,-292.6667) and (430.3333,-331.3333) ..
    (409.0000,-384.0000) -- (386.0000,-441.0000) .. controls (347.3333,-537.0000)
    and (318.0000,-603.0000) .. (298.0000,-639.0000) .. controls
    (248.0000,-727.0000) and (198.0000,-773.6667) .. (148.0000,-779.0000) ..
    controls (202.0000,-828.3333) and (249.0000,-853.0000) .. (289.0000,-853.0000)
    .. controls (343.6667,-853.0000) and (404.3333,-778.6667) ..
    (471.0000,-630.0000) -- (507.0000,-550.0000) .. controls (755.6667,-996.0000)
    and (1074.6667,-1335.3333) .. (1464.0000,-1568.0000) -- (1494.0000,-1516.0000)
    -- cycle;
\end{scope}

\end{tikzpicture}} and \resizebox{!}{\fontcharht\font`M}{\begin{tikzpicture}[y=0.80pt, x=0.80pt, yscale=-1.000000, xscale=1.000000, inner sep=0pt, outer sep=0pt]
\begin{scope}[shift={(100.0,1731.0)},nonzero rule]
  \path[draw=.,fill=.,line width=1.600pt] (1271.0000,-1280.0000) ..
    controls (1271.0000,-1272.0000) and (1263.6667,-1261.0000) ..
    (1249.0000,-1247.0000) .. controls (1237.0000,-1235.6667) and
    (1213.6667,-1207.6667) .. (1179.0000,-1163.0000) -- (866.0000,-754.0000) --
    (929.0000,-627.0000) -- (1089.0000,-285.0000) .. controls
    (1083.0000,-271.0000) and (1073.6667,-261.3333) .. (1061.0000,-256.0000) --
    (1061.0000,-240.0000) .. controls (1060.3333,-239.3333) and
    (1052.3333,-234.3333) .. (1037.0000,-225.0000) .. controls
    (1025.6667,-217.6667) and (1020.0000,-211.6667) .. (1020.0000,-207.0000) ..
    controls (1020.0000,-201.0000) and (1023.3333,-191.3333) ..
    (1030.0000,-178.0000) .. controls (1028.0000,-166.0000) and
    (1019.3333,-154.3333) .. (1004.0000,-143.0000) .. controls
    (988.6667,-132.3333) and (974.6667,-127.0000) .. (962.0000,-127.0000) ..
    controls (948.6667,-127.0000) and (922.6667,-157.0000) .. (884.0000,-217.0000)
    -- (692.0000,-514.0000) -- (286.0000,115.0000) .. controls (264.6667,131.0000)
    and (229.3333,139.0000) .. (180.0000,139.0000) .. controls (168.6667,139.0000)
    and (154.3333,125.3333) .. (137.0000,98.0000) .. controls (116.3333,65.3333)
    and (106.0000,44.3333) .. (106.0000,35.0000) .. controls (106.0000,27.0000)
    and (108.0000,16.0000) .. (112.0000,2.0000) .. controls (118.0000,-18.6667)
    and (117.3333,-32.0000) .. (110.0000,-38.0000) .. controls (103.3333,-44.6667)
    and (100.0000,-49.6667) .. (100.0000,-53.0000) .. controls (100.0000,-64.3333)
    and (116.3333,-98.0000) .. (149.0000,-154.0000) -- (198.0000,-238.0000) ..
    controls (250.6667,-328.0000) and (367.6667,-482.6667) .. (549.0000,-702.0000)
    -- (573.0000,-731.0000) -- (422.0000,-1149.0000) .. controls
    (406.6667,-1191.6667) and (399.0000,-1215.0000) .. (399.0000,-1219.0000) ..
    controls (399.0000,-1225.0000) and (408.0000,-1239.3333) ..
    (426.0000,-1262.0000) -- (440.0000,-1274.0000) .. controls
    (444.6667,-1278.0000) and (449.3333,-1280.0000) .. (454.0000,-1280.0000) ..
    controls (461.3333,-1280.0000) and (472.3333,-1274.0000) ..
    (487.0000,-1262.0000) .. controls (495.0000,-1268.6667) and
    (506.0000,-1272.0000) .. (520.0000,-1272.0000) .. controls
    (524.0000,-1274.0000) and (530.0000,-1278.6667) .. (538.0000,-1286.0000) ..
    controls (544.6667,-1297.3333) and (552.3333,-1303.0000) ..
    (561.0000,-1303.0000) .. controls (573.6667,-1303.0000) and
    (586.0000,-1290.6667) .. (598.0000,-1266.0000) -- (753.0000,-958.0000) --
    (805.0000,-1022.0000) -- (905.0000,-1137.0000) .. controls
    (1029.6667,-1280.3333) and (1095.0000,-1352.0000) .. (1101.0000,-1352.0000) ..
    controls (1117.6667,-1352.0000) and (1141.6667,-1339.0000) ..
    (1173.0000,-1313.0000) -- (1196.0000,-1333.0000) .. controls
    (1205.3333,-1347.0000) and (1212.6667,-1354.0000) .. (1218.0000,-1354.0000) ..
    controls (1221.3333,-1354.0000) and (1225.3333,-1348.3333) ..
    (1230.0000,-1337.0000) .. controls (1232.6667,-1331.0000) and
    (1240.6667,-1320.6667) .. (1254.0000,-1306.0000) .. controls
    (1265.3333,-1294.0000) and (1271.0000,-1285.3333) .. (1271.0000,-1280.0000) --
    cycle;
\end{scope}

\end{tikzpicture}} to indicate whether contain the corresponding block.

\subsection{Qualitative Comparisons}
\noindent {\bf{Baseline comparisons.}} \quad Fig. \ref{fig1} shows the visualization results of the qualitative comparison. Three rows in the top left corner show the experimental results on the \emph{AFHQ50} datasets, we can find that SinGAN, ConSinGAN, PetsGAN, and SinDiffusion are less effective in terms of global structure consistency due to the lack of global information representation constraints. Moreover, ExSinGAN is followed only by our TcGAN method. Three rows in the bottom left corner show the experimental results on the \emph{CelebA50} dataset, it is obvious that all methods except TcGAN have problems such as content distortion and unreasonable global structure. The right side show the experimental results on the \emph{Places50} dataset, where we can find that TcGAN achieves consistency in the global structure while achieving the best visualization results for local features. The other state-of-the-art methods are consistent in terms of local layout, but they all show blurring and texture roughness in terms of local features. From the above comparison experiments, we can observe that the proposed method TcGAN achieves the consistency of global structure as well as the superiority of local features based on the global information representation.

\noindent {\bf{Presentation of the results of different stages.}} \quad As shown in Fig. \ref{fig5}, we have visualized the images generated by each method at each stage of the training process. According to the displayed results, we can observe that TcGAN generates images with global structure and visual consistency at the first stage. At the same time,  the texture information of the generated images becomes more visible as the local network deepens, which cannot be achieved by other baselines. According to the images generated by SinGAN during the first, second, and third stages, it can be found that the global structure information is not reasonable, because the generation results in the third stage are not as good as those generated by ExSinGAN and TcGAN in the first stage. In addition, the global structure keeps changing as the generator is trained, but still does not achieve visual consistency (e.g., the second and third rows). The output of ConSinGAN in the third stage can be identical to TcGAN in the first stage. ExSinGAN and PetsGAN accomplish global structure consistency in the first stage, but cannot effectively maintain local texture information during the later training (e.g., the second row).

\noindent {\bf{Study of the number of Transformer.}} \quad As shown in Fig. \ref{fig4}, we explore the impact of the number of global networks on TcGAN during the training period. It can be found that at $ n $=1, TcGAN is able to achieve the complete global structure of the training image. Meanwhile, as the number of $n$ increases, the global structure of the output image remains similar to that of $ n $=1. In addition, as the number of $ n $ increases, the number of learnable parameters becomes more leading to an increase in training time. In summary, from both time efficiency as well as experimental results we can conclude that TcGAN is capable of performing image generation tasks within the period of $ n $=1.

\noindent {\bf{Random Sample.}} \quad Fig. \ref{fig0} and Fig. \ref{fig10} present examples of random sample images generated by TcGAN, where Fig. \ref{fig0} exhibits more examples but only one result per example, while Fig. \ref{fig10} demonstrates fewer examples but multiple results per example. It can be seen that our method can not only generate realistic random image samples and also has the performance of preserving the patch distribution of training images as well as describing new structures.

\noindent {\bf{Comparison experiments with different number of iterations.}} \quad As shown in Fig. \ref{fig3}, we evaluate TcGAN against other state-of-the-art methods on three datasets with different numbers of iterations. We can find that SinGAN improves the quality of generated images as the number of iterations increases in all cases. However, in the first row of the \emph{AFHQ50} dataset experiment, the local features of the generated images appear distorted; in the second and third rows of the \emph{Places50} dataset experiment, the global structure, as well as the texture, are not reasonable. ConSinGAN experiments on the three datasets at $iter$=500 and 1000 not only did not achieve global consistency but also local features were not extracted effectively. At $iter$=2000, only the last row of the \emph{CelebA} dataset achieves general results. The global structure of TcGAN is not only consistent at $iter$=500 for the three datasets, but also the local features become more refined as the image quality increases with the number of iterations, which reflects the effectiveness of the global network.

\noindent {\bf{Comparison experiments with different scaling formulations.}} \quad As shown in Fig. \ref{fig13}, we evaluate TcGAN with different scaling formulations on five datasets. The columns TcGAN+S denote TcGAN with the scaling method of SinGAN, TcGAN+C indicate TcGAN with the scaling method of ConSinGAN and TcGAN is the proposed scaling method. We can observe from Fig. \ref{fig13}(a) that both TcGAN+S and TcGAN+C demonstrate global inconsistency as well as a phantom in the low-resolution group, and become worse in the high-resolution group as shown in Fig. \ref{fig13}(b). 

\noindent {\bf{Super-Resolution.}} \quad As shown in Fig. \ref{fig7}, we compare the experimental results of SinGAN, ConSinGAN, and TcGAN on super-resolution tasks. SinGAN generates super-resolution images by increasing the minimum size after scaling and repeating the residual connect input operation, which achieves visual consistency but suffers from local details and blurring (e.g., the first and second rows of the second column). PetsGAN achieves higher semantic information for super-resolution images but ignores the importance of global structure leading to overlapping, phantom, etc. (e.g., the first and third rows of the third column). Compared to the baselines, TcGAN achieves excellent performance on both datasets.

\begin{figure*}[ht]
	\centering
	\includegraphics[width=7.16in]{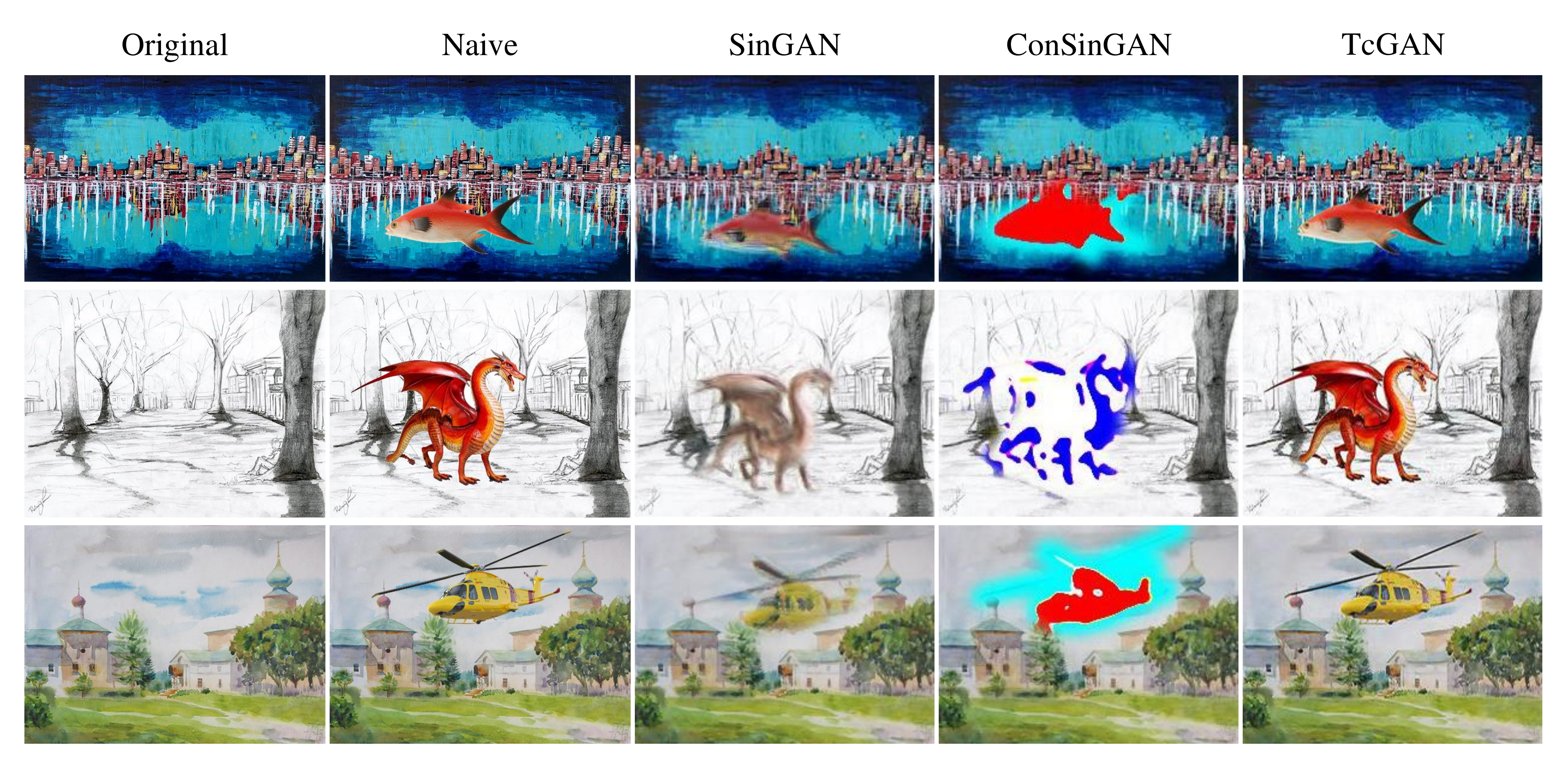}
	\caption{{\bf{Harmonization.}} Reasonably blend an object image on a background image. The first two columns show the image with the original size (the maximum size is 600), while the last three columns of the comparison methods show the size of the generated image (the maximum size is 250).}
	\label{fig8}
\end{figure*}

\noindent {\bf{Complicated Background.}} \quad As shown in Fig. \ref{fig2}, we conduct comparative experiments on the $ Places50 $ dataset by randomly sampling complex structure images with strong granularity rationality to study the robustness and generalization of different methods. The generation of complex structured images is extremely demanding in terms of semantic-aware information processing of local details. We can observe that SinGAN, ConSinGAN, ExSinGAN, and PetsGAN all lose the content of the training image and also lack global structure rationality. On the contrary, TcGAN achieves the best performance because it can preserve both the complete layout and structure of the training image as well as the realistic local textures and details.

\noindent {\bf{Image Harmonization.}} As shown in Fig. \ref{fig8}, we evaluate SinGAN, ConSinGAN, and TcGAN in other image processing application scenarios. The high fidelity of the SinGAN generated image is lower, as shown by the inclusion of only a small amount of color of the subject image and the blurring phenomenon. ConSinGAN is trained (without any fine-tuning) to retain only the outline of the appearance of the subject image, completely losing the structure and local detail information. Compared to the baseline, TcGAN still achieves better results in terms of reality and visual perception, thanks to its complete preservation of the consistent structure and color of the subject image and the more obvious texture information.

\subsection{Quantitative Results}
We quantitatively compare the proposed TcGAN with the adopted methods in terms of quality, diversity, similarity, and training time of the method. Similar to PetsGAN\cite{zhang2022petsgan35}, we also use Single Image Fréchet Inception Distance (SIFID)\cite{shaham2019singan9} and Learned Perceptual Image Patch Similarity (LPIPS)\cite{zhang2018perceptual53} to measure the quality and diversity of the generated images, respectively. In addition, we also used Structural SIMilarity (SSIM)\cite{wang2004image54} to measure the similarity. 

\noindent {\bf{SIFID.}} \quad The SIFID which represents the Fréchet Inception Distance (FID)\cite{heusel2017gans55} of a single image view measures the FID distance between two images from the depth feature, and the closer this distance is, the better the generated method is, i.e., the images are high in clarity and rich in diversity. 

\noindent {\bf{LPIPS.}} \quad The LPIPS metric measures the distance between two image patches in terms of human visual perception of textured images. A higher LPIPS value means that they are more different from each other, and lower means more similar. 

\noindent {\bf{SSIM.}} \quad The SSIM which measures the quality of the training image and the generated image in terms of brightness, contrast, and structure is used to evaluate the similarity of the two images. The closer the SSIM value is to 1, the better the quality; and the closer it is to 0, the worse the quality.
\begin{table*}[ht]
	\begin{center}
		\caption{Evaluations of SIFID, LPIPS and SSIM on the AFHQ50, Places50 and CelebA50. Mean is the average results on the three datasets. $\downarrow$ indicates that
			the lower the better while $\uparrow$ indicates the higher the better.}
		\label{tab2}
		\begin{tabular}{|c|ccc|ccc|ccc|ccc|}
			\hline
			{\bf{Datasets}}&\multicolumn{3}{c|}{\emph{AFHQ50}}&\multicolumn{3}{c|}{\emph{Places50}}&\multicolumn{3}{c|}{\emph{CelebA50}} & \multicolumn{3}{c|}{Mean}\\
			\hline
			{\bf{Matric}}& \bf{SIFID}$\downarrow$ & \bf{LPIPS}$\downarrow$ &  \bf{SSIM}$\uparrow$ & \bf{SIFID}$\downarrow$ & \bf{LPIPS}$\downarrow$ &  \bf{SSIM}$\uparrow$ & \bf{SIFID}$\downarrow$ & \bf{LPIPS}$\downarrow$ &  \bf{SSIM}$\uparrow$  & \bf{SIFID}$\downarrow$ & \bf{LPIPS}$\downarrow$ &  \bf{SSIM}$\uparrow$ \\
			\hline
			SinGAN & 8.2110	& 0.5315 & 0.3763 & 12.1633 & 0.5200 & 0.2949 & 6.0961 & 0.4355 & 0.4314 & 8.8235 & 0.4957 & 0.3675 \\ 
			\hline
			ConSinGAN & 0.0499 & 0.3375	& 0.5186 & 0.0755 & 0.2365 & 0.4364 & 0.0343 & 0.2025 & 0.5901 & 0.0532 & 0.2588 & 0.5150 \\
			\hline
			ExSinGAN & 0.0364 & 0.1275 & 0.7807 & 0.0803 & 0.1745 & 0.7215 & 6.0927	& 0.1005 & 0.7806 & 2.0698 & 0.1342 & 0.7609 \\
			\hline
			PetsGAN & 0.1472 & 0.3660 & 0.4871 & 0.1079 & 0.2835	& 0.4976 & 0.0464 & 0.1985 & 0.6398 & 0.1005 & 0.2827 & 0.5415 \\
			\hline
			SinDiffusion & 8.2295 & 0.5520 & 0.3031 & 12.1716 & 0.5510 & 0.3160 & 6.0599 & 0.4835 & 0.3352 & 8.8203 & 0.5288 & 0.3181 \\
								
			\hline
			\hline
			TcGAN & \bf{0.0220} & \bf{0.0745}	& \bf{0.8163} & \bf{0.0117} & \bf{0.0560} & \bf{0.8449}	& \bf{0.0242} & \bf{0.0615} & \bf{0.8882} & \bf{0.0193} & \bf{0.0640} & \bf{0.8498} \\
			\hline 
		\end{tabular}
	\end{center}
\end{table*}

\begin{figure}[!t]
	\centering
	\includegraphics[width=3.5in]{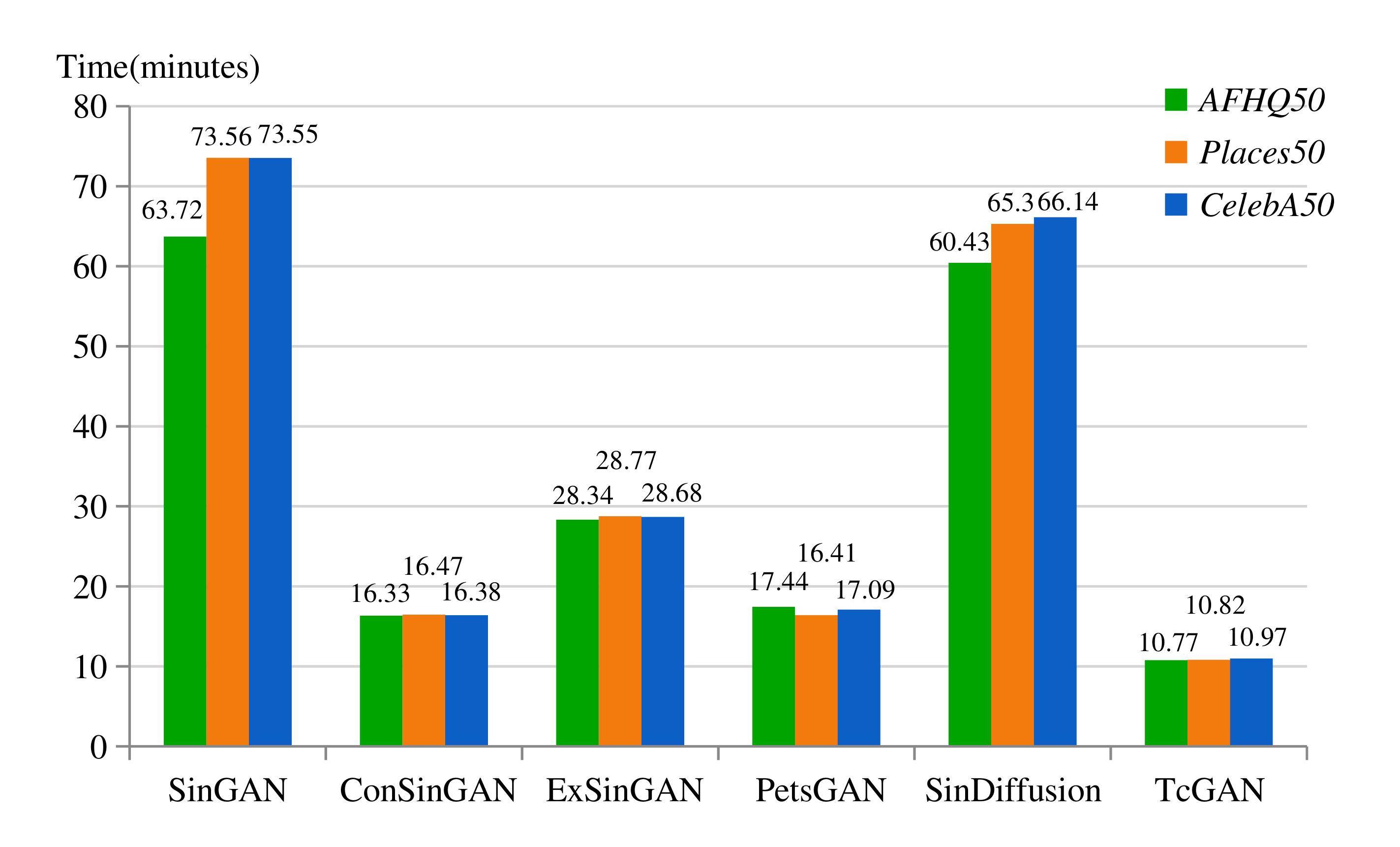}
	\caption{The training time of different OSG methods on three datasets.}
	\label{fig6}
\end{figure}

As shown in Table \ref{tab2}, TcGAN achieves optimal values in terms of SIFID, LPIPS, SSIM, and average metrics on all datasets compared with other methods, indicating that TcGAN is capable of generating images with realistic texture capabilities together with global structures with natural layouts. The same conclusion as the visualization result is obtained, which is in accordance with the result obtained from the visualization. In addition, as shown in Fig. \ref{fig6}, our TcGAN method also has a significant advantage in training efficiency. The average training time is about 10 minutes on the same experiment environment, which is about 1.6 times faster than ConSinGAN and PetsGAN, about 2.8 times faster than ExSinGAN, and about 6.5 times faster than SinGAN and SinDiffusion. This demonstrates that TcGAN can generate training images with more complete global structure using less training time compared to other state-of-the-art methods.
\section{Conclusion}
We propose a novel method TcGAN that enables arbitrary image generation. The proposed global network by using a individual structure-preserved transformer focuses on solving the problems such as illusions and overlaps that appear in previous OSG methods. In addition, we design a new scaling method to extend our proposed TcGAN to achieve super-resolution tasks. The method has scale-invariance during scaling allow the generated super-resolution images to retain the original input semantic-aware information without any additional labeling. Extensive experiments demonstrate that TcGAN can not only generate arbitrary images but train and perform OSG tasks at a much faster rate. Furthermore, we have extended TcGAN to image harmonization. In the future, we will consider further extending our method so that the generated images are diverse while maintaining global structural consistency.

\end{document}